\documentclass{article}

    \PassOptionsToPackage{numbers, compress}{natbib}

    \usepackage[final]{neurips_2021}

\usepackage[utf8]{inputenc} 
\usepackage[T1]{fontenc}    
\usepackage{hyperref}       
\usepackage{url}            
\usepackage{booktabs}       
\usepackage{amsfonts}       
\usepackage{nicefrac}       
\usepackage{microtype}      
\usepackage{xcolor}         
\usepackage{subcaption}
\usepackage{epsfig}

\usepackage{xspace}
\usepackage{amsmath}
\usepackage{amssymb}
\usepackage{array}
\usepackage{wrapfig}

\def\*#1{\mathbf{#1}}

\makeatletter
\DeclareRobustCommand\onedot{\futurelet\@let@token\@onedot}
\def\@onedot{\ifx\@let@token.\else.\null\fi\xspace}

\def\ie{\emph{i.e}\onedot}

\def\wrt{\emph{w.r.t}\onedot} 

\makeatother
\usepackage{multirow}

\newcolumntype{C}[1]{>{\centering\let\newline\\\arraybackslash}p{#1}}

\title{On the Importance of Gradients for Detecting Distributional Shifts in the Wild}

\author{%
  Rui Huang \\
  Department of Computer Sciences\\
  University of Wisconsin-Madison\\
  \texttt{huangrui@cs.wisc.edu} \\
  \And 
  Andrew Geng \\
  Department of Computer Sciences\thanks{Work done while A.G was working as an undergraduate research assistant with Li's lab.}\\
  University of Wisconsin-Madison\\
  \texttt{ageng@wisc.edu} \\
  \And
  Yixuan Li \\
  Department of Computer Sciences\\
  University of Wisconsin-Madison\\
  \texttt{sharonli@cs.wisc.edu} \\
}

\begin{document}

\maketitle

\begin{abstract}
Detecting out-of-distribution (OOD) data has become a critical component in ensuring the safe deployment of machine learning models in the real world. 
Existing OOD detection approaches primarily rely on the output or feature space for deriving OOD scores, while largely overlooking information from the \emph{gradient space}.
In this paper, we present \texttt{GradNorm}, a simple and effective approach for detecting OOD inputs by utilizing information extracted from the gradient space. \texttt{GradNorm} directly employs the vector norm of gradients, backpropagated from the KL divergence between the softmax output and a uniform probability distribution. Our key idea is that the magnitude of gradients is higher for in-distribution (ID) data than that for OOD data, making it informative for OOD detection. \texttt{GradNorm} demonstrates superior performance, reducing the average FPR95 by up to {16.33\%} compared to the previous best method. Code and data available: \url{https://github.com/deeplearning-wisc/gradnorm\_ood}.
\end{abstract}

\vspace{-\baselineskip}
\section{Introduction}

When deploying machine learning models in the real world, there is an increasingly important question to ask: \emph{``Is the model making a faithful prediction for something it was trained on, or is the model making an unreliable prediction for something it has not been exposed to during training?''} We want models that are not only accurate on their familiar data distribution, but also aware of uncertainty outside the training distribution. 
This gives rise to the importance of out-of-distribution (OOD) detection, which determines whether an input is in-distribution (ID) or OOD. As of recently a plethora of literature has emerged to address the problem of OOD uncertainty estimation~\cite{chen2021robustifying, hendrycks2016baseline, hsu2020generalized, huang2021mos, lakshminarayanan2017simple, lee2018simple, liang2018enhancing, lin2021mood, liu2020energy,  mohseni2020self, nalisnick2018deep}.

The main challenge in OOD uncertainty estimation stems from the fact that modern deep neural networks can easily produce overconfident predictions on OOD inputs~\cite{nguyen2015deep}. This phenomenon makes the separation between ID and OOD data a non-trivial task. Much of the prior work focused on deriving OOD uncertainty measurements from the activation space of the neural network, e.g., using model output~\cite{hendrycks2016baseline, hsu2020generalized,lakshminarayanan2017simple,liang2018enhancing, liu2020energy} or feature representations~\cite{lee2018simple}. Yet, this leaves an alternative space---model {parameter and its \emph{gradient space}}---largely unexplored. 
Will a model react to ID and OOD inputs differently in its gradient space, and if so, can we discover distinctive signatures to separate ID and OOD data from gradients?

\begin{wrapfigure}{r}{0.45\textwidth}
\vspace{-1cm}
    \includegraphics[width=0.45\textwidth]{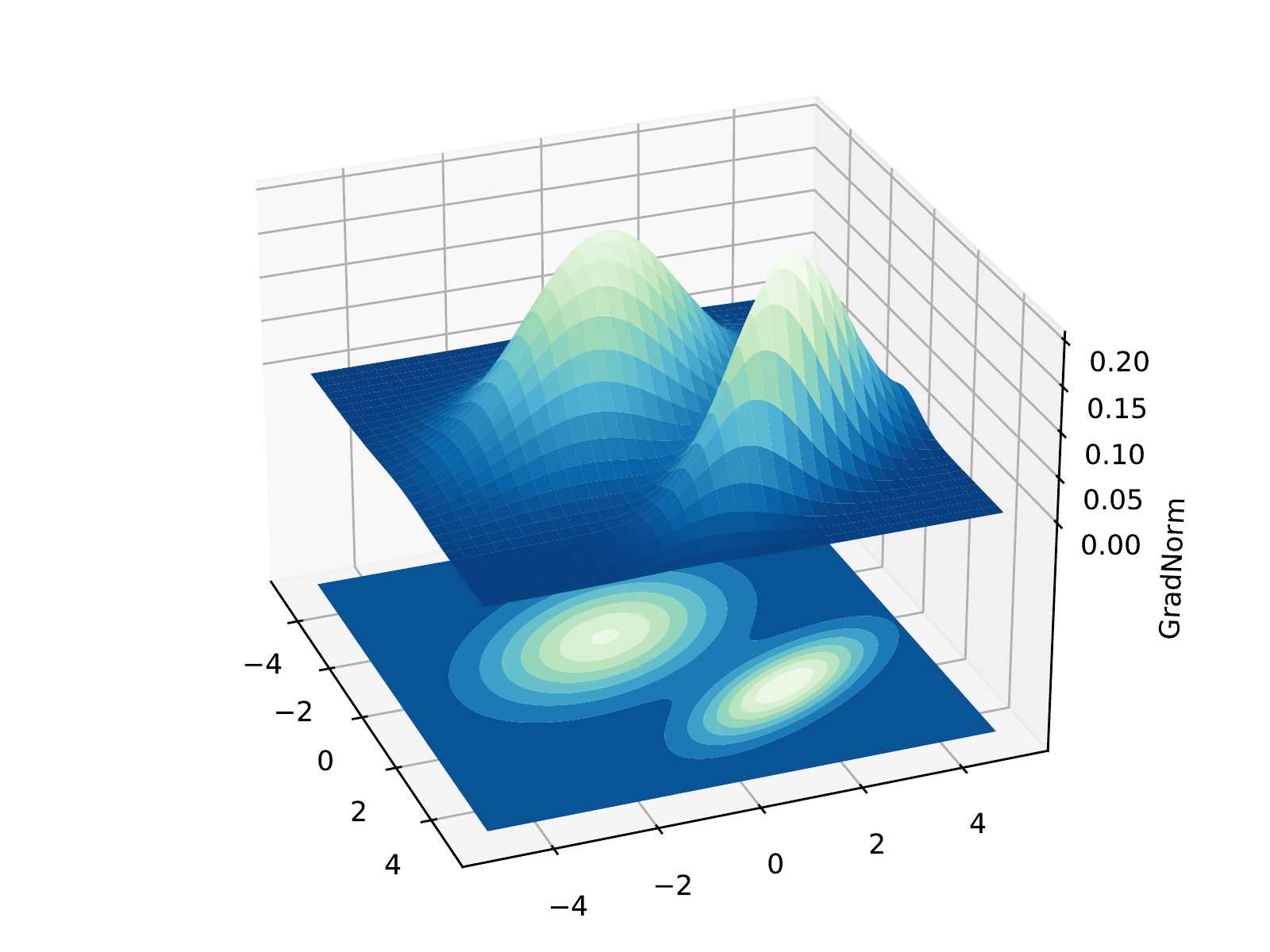}
        \vspace{-0.5cm}
    \caption{\small{An example of two-dimensional input space. Input data is depicted in the $xy$-plane, while gradient norm for each input is depicted in the $z$-dimension. The magnitude of gradients is higher for ID data ({\color{green}light green}) than that for OOD data ({\color{blue}deep blue}).}}
    \label{fig:teaser}
    \vspace{-1.5cm}
\end{wrapfigure}

In this paper, we tackle this key question by exploring and exploiting the richness of the {gradient space}, ultimately showing that {gradients} carry surprisingly useful signals for OOD detection. Formally, we present \texttt{GradNorm}, a simple and effective approach for detecting OOD inputs by utilizing gradient extracted from a pre-trained neural network. Specifically, \texttt{GradNorm} employs the vector norm of gradients directly as an OOD scoring function. Gradients are backpropagated from the Kullback-Leibler (KL) divergence~\cite{kullback1951information} between the softmax output and a uniform distribution. ID data is expected to have larger KL divergence because the prediction tends to concentrate on one of the ground-truth classes and  is therefore  less uniformly distributed. 
As depicted in Figure~\ref{fig:teaser}, our key idea is that the gradient norm of the KL divergence is higher for ID data than that for OOD data, making it informative for OOD uncertainty estimation. 

We provide both empirical and theoretical insights, demonstrating the superiority of \texttt{GradNorm} over both output-based and feature-based methods. Empirically, we establish superior performance on a large-scale ImageNet benchmark, as well as a suite of common OOD detection benchmarks. 
\texttt{GradNorm} outperforms the previous best method by a large margin, with up to {16.33}\% reduction in false-positive rate (FPR95). Theoretically, we show that \texttt{GradNorm} captures the \emph{joint information} between the feature and the output space. The joint information results in an overall stronger separability than using either feature or output space alone. 

 

Our \textbf{key results and contributions} are summarized as follows. 
\begin{itemize}
  \item We propose \texttt{GradNorm}, a simple and effective gradient-based OOD uncertainty estimation method, which is both label-agnostic (no label required for backpropagation) and OOD-agnostic (no outlier data required). \texttt{GradNorm} reduces the average FPR95 by {16.33\%} compared to the current best method.
  \item We perform comprehensive analyses that improve understandings of the gradient-based method under (1) different network architectures, (2) gradient norms extracted at varying depths, (3) different loss functions for backpropagation, and (4) different vector norms for aggregating gradients. 
  \item We perform a mathematical analysis of \texttt{GradNorm} and show that it can be decomposed into two terms, jointly characterizing information from both feature and output space, which demonstrates superiority. 
\end{itemize}

\vspace{-0.2cm}
\section{Preliminaries}
\label{sec:background}
\vspace{-0.2cm}
We start by recalling the general setting of the supervised learning problem. We denote by $\mathcal{X}=\mathbb{R}^d$ the input space and $\mathcal{Y}=\{1,2,..., C\}$ the output space. A learner is given access to a set of training data $D=\{(\*x_i,y_i)\}_{i=1}^N$ drawn from an unknown joint data distribution $P$ defined on $\mathcal{X}\times \mathcal{Y}$. 
A neural network $f(\*x;\theta): \mathcal{X} \to \mathbb{R}^C$ minimizes the empirical risk: 
\begin{align*}
R_\mathcal{L}(f)=\mathbb{E}_{D}(\mathcal{L}_\text{CE}(f(\*x;\theta),y)),
\end{align*}
where $\theta$ is the parameters of the network, and $\mathcal{L}_\text{CE}$ is the commonly used cross-entropy loss:
\begin{align}
    \mathcal{L}_\text{CE} (f(\*x),y) &= - \log{ \frac{e^{{f_{y}(\*x)} / T}}{\sum_{c=1}^C e^{{f_{c}(\*x)} / T}}}.
\end{align}
Specifically, $f_y(\*x)$ denotes the $y$-th element of $f(\*x)$ corresponding to the ground-truth label $y$, and $T$ is the temperature.

\vspace{-0.2cm}
\paragraph{Problem statement}
Out-of-distribution (OOD) detection can be formulated as a binary classification problem. In practice, OOD is often defined by a distribution that simulates unknowns encountered during deployment time, such as samples from an irrelevant distribution whose label set has no intersection with $\mathcal{Y}$ and \emph{therefore should not be predicted by the model}.  Given a classifier $f$ learned on training samples from in-distribution $P$, the goal is to design a binary function estimator,
\begin{equation*}
    g(\*x) = 
    \begin{cases}
    \text{in}, &\text{if}\ S(\*x) \geq \gamma \\
    \text{out}, &\text{if}\ S(\*x) < \gamma,
    \end{cases}
\end{equation*}
that classifies whether a sample $\*x\in \mathcal{X}$ is from $P$ or not. $\gamma$ is commonly chosen so that a high fraction (e.g., 95\%) of ID data is correctly classified. The key challenge is to derive a scoring function $S(\*x)$ that captures OOD uncertainty. Previous approaches have primarily relied on the model's output or features for OOD uncertainty estimation. Instead, our approach seeks to compute $S(\*x)$ based on the information extracted from the \emph{gradient space}, which we describe in the next section.

\section{Gradient-based OOD Detection}
\label{sec:method}

In this section, we describe our  method \texttt{GradNorm}. We start by introducing the loss function for backpropagation and then describe how to leverage the gradient norm for OOD uncertainty estimation.

We calculate gradients \wrt each parameter by backpropagating the Kullback-Leibler (KL) divergence~\cite{kullback1951information} between the softmax output and a uniform distribution. Formally, KL divergence quantifies how close a model-predicted distribution $q=\{q_i\}$ is to a reference probability distribution $p=\{p_i\}$,
\begin{equation}
    D_\text{KL} (p~||~q) = \sum_i p_i \log \frac{p_i}{q_i} = -\sum_i p_i \log q_i + \sum_i p_i \log p_i  = H(p,q) - H(p).
\end{equation}

In particular, we set the reference distribution to be uniform $\*u= [1/C, 1/C,...,1/C] \in \mathbb{R}^C$. The predictive probability distribution is the softmax output. Our KL divergence for backpropagation can be written as:
\vspace{-0.2cm}
\begin{equation}
\label{eq:kl}
    D_\text{KL}(\*u~||~\text{softmax}(f(\*x))  = - \frac{1}{C}\sum_{c=1}^C \log{ \frac{e^{{f_c(\*x)} / T}}{\sum_{j=1}^C e^{{f_{j}(\*x)} / T}}} - H(\*u),
\end{equation}
where the first term is the cross-entropy loss between the softmax output and a uniform vector $\*u$, and the second term $H(\*u)$ is a constant. The KL divergence measures how much the predictive distribution is away from the uniform distribution. Intuitively, ID data is expected to have larger KL divergence because the prediction tends to concentrate on the ground-truth class and is thus distributed less uniformly. 

\textbf{{GradNorm} as OOD score} For a given parameter $w$, the gradient of the above KL divergence is:
\begin{align}
    \frac{\partial D_\text{KL}(\*u~||~\text{softmax}(f(\*x))}{\partial w} = \frac{1}{C}\sum_{i=1}^C \frac{\partial \mathcal{L}_\text{CE}(f(\*x),i)}{\partial w},
\end{align}
where $w$ is a component of network parameter $\theta$. Notice that the gradient of the entropy term is $0$, \ie $\partial H(\*u) / \partial w=0$. In other words, the gradient of KL divergence is equivalent to averaging the derivative of the categorical cross-entropy loss for \emph{all} labels.

We now define the OOD score via a vector norm of gradients of the selected parameters:
\begin{equation}
\label{eq:score}
    S(\*x) = \lVert\frac{\partial D_\text{KL}(\*u~\lVert~\text{softmax}(f(\*x))}{\partial \*w}  \rVert_p,
\end{equation}
where $\lVert \cdot \rVert_p$ denotes $L_p$-norm and $\*w$ is the set of   parameters in vector form\footnote{We concatenate all selected parameters into a single vector regardless of the original shapes of the parameters.}. We term our method~\texttt{GradNorm}, short for gradient norm. In practice, ~\texttt{GradNorm} can be conveniently implemented by calculating the cross-entropy loss between the predicted softmax probability and a uniform vector as the target. We will discuss the choices and impacts of the selected parameter set $\*w$ in Section~\ref{sec:ablations}. 




\textbf{Rationale of {GradNorm}} Our operating hypothesis is that using the KL divergence for backpropagation, the gradient norm is higher for ID data than that for OOD data. As we show in Section~\ref{sec:ablations}, using the gradient norm of the KL divergence is more effective than using the KL divergence directly.
Moreover, \texttt{GradNorm} derived from the KL divergence with a uniform target offers two advantages over gradient norms derived from the standard cross-entropy loss.
\vspace{-0.2cm}
\begin{itemize}
\itemsep0em
    \item First, our method is \emph{label-agnostic} and does not require any ground-truth label. It can be flexibly used during inference time when the label is unavailable for either ID or OOD data. 
    \item  Second, it captures the \emph{uncertainty across all categories}, providing more information for OOD detection. We will provide empirical evidence to show the importance of utilizing all labels in Section~\ref{sec:ablations}.
\end{itemize}

\vspace{-0.3cm}

\section{Experiments}
\label{sec:experiments}
\vspace{-0.3cm}
In this section, we evaluate \texttt{GradNorm} on a large-scale OOD detection benchmark with ImageNet-1k as in-distribution dataset~\cite{huang2021mos}. We describe experimental setup in Section~\ref{sec:exp_setup} and demonstrate the superior performance of \texttt{GradNorm} over existing approaches in Section~\ref{sec:ablations}, followed by extensive ablations and analyses that improve the understandings of our approach. 

\vspace{-0.2cm}
\subsection{Experimental Setup}
\label{sec:exp_setup}
\vspace{-0.2cm}
\textbf{Dataset} 
We use the large-scale ImageNet OOD detection benchmark proposed by \citeauthor{huang2021mos}~\cite{huang2021mos}.
ImageNet benchmark is not only more realistic (with higher resolution images) but also more challenging (with a larger label space of 1,000 categories). 
We evaluate on four OOD test datasets, which are from subsets of \texttt{iNaturalist}~\cite{van2018inaturalist}, \texttt{SUN}~\cite{xiao2010sun}, \texttt{Places}~\cite{zhou2017places}, and \texttt{Textures}~\cite{cimpoi2014describing}, with non-overlapping categories \wrt ImageNet-1k (see Appendix~\ref{app:dataset} for detail). The evaluations span a diverse range of domains including fine-grained images, scene images, and textural images. We further evaluate on CIFAR benchmarks that are routinely used in literature (see Appendix~\ref{app:cifar}).
\textbf{Model and hyperparameters} 
We use Google BiT-S models\footnote{\url{https://github.com/google-research/big_transfer}}~\cite{kolesnikov2020big} pre-trained on ImageNet-1k with a ResNetv2-101 architecture~\cite{he2016identity}. We report performance on an alternative architecture, DenseNet-121~\cite{huang2017densely}, in Section~\ref{sec:ablations}. 
Additionally, we use $L_1$-norm-based OOD scores as the default and explore the effect of other $L_p$-norms in Section~\ref{sec:ablations}. The temperature parameter $T$ is set to be 1 unless specified otherwise, and we explore the effect of different temperatures in Section~\ref{sec:ablations}. At test time, all images are resized to 480 $\times$ 480.



\begin{table*}[t]
\centering
\scriptsize{

\begin{tabular}{c|l|C{0.03\textwidth}C{0.045\textwidth}|C{0.03\textwidth}C{0.045\textwidth}|C{0.03\textwidth}C{0.045\textwidth}|C{0.03\textwidth}C{0.045\textwidth}|C{0.03\textwidth}C{0.045\textwidth}}
\toprule
\multirow{3}{*}{\textbf{\begin{tabular}[c]{@{}c@{}}Method\\ Space\end{tabular}}} & \multicolumn{1}{c|}{\multirow{3}{*}{\textbf{Method}}} & \multicolumn{2}{c|}{\textbf{iNaturalist}}    & \multicolumn{2}{c|}{\textbf{SUN}}            & \multicolumn{2}{c|}{\textbf{Places}}         & \multicolumn{2}{c|}{\textbf{Textures}}       & \multicolumn{2}{c}{\textbf{Average}}        \\ \cline{3-12} 
                                                                                      & \multicolumn{1}{c|}{}                                 & \tiny{FPR95}                & \tiny{AUROC}                 & \tiny{FPR95}                & \tiny{AUROC}              & \tiny{FPR95}                & \tiny{AUROC}                 & \tiny{FPR95}                & \tiny{AUROC}                 & \tiny{FPR95}                & \tiny{AUROC}               \\
                                                                                      & \multicolumn{1}{c|}{}                                 & \multicolumn{1}{c}{$\downarrow$} & \multicolumn{1}{c|}{$\uparrow$} & \multicolumn{1}{c}{$\downarrow$} & \multicolumn{1}{c|}{$\uparrow$} & \multicolumn{1}{c}{$\downarrow$} & \multicolumn{1}{c|}{$\uparrow$} & \multicolumn{1}{c}{$\downarrow$} & \multicolumn{1}{c|}{$\uparrow$} & \multicolumn{1}{c}{$\downarrow$} & \multicolumn{1}{c}{$\uparrow$}  \\ \midrule
                                
\multirow{3}{*}{Output}                                                         & MSP\tiny{~\cite{hendrycks2016baseline}}                                                   & 63.69                & 87.59                 & 79.98                & 78.34                 & 81.44                & 76.76                 & 82.73                & 74.45                 & 76.96                & 79.29                \\
                                                                                      & ODIN\tiny{~\cite{liang2018enhancing}}                                                  & 62.69                & 89.36                 & 71.67                & 83.92                 & 76.27                & 80.67                 & 81.31                & 76.30                 & 72.99                & 82.56                \\
                                                                                      & Energy\tiny{~\cite{liu2020energy}}                                                & 64.91                & 88.48                 & 65.33                & 85.32                 & 73.02                & 81.37                 & 80.87                & 75.79                 & 71.03                & 82.74                \\
                                                                                      \midrule
Feature                                                                         & Mahalanobis\tiny{~\cite{lee2018simple}}                                            & 96.34                & 46.33                 & 88.43                & 65.20                 & 89.75                & 64.46                 & \textbf{52.23}                & 72.10                 & 81.69                & 62.02                \\ \midrule
\multirow{1}{*}{Gradient}                    
                                                                                      & \textbf{GradNorm (ours)}                              & \textbf{50.03}       & \textbf{90.33}        & \textbf{46.48}       & \textbf{89.03}        & \textbf{60.86}       & \textbf{84.82}        & 61.42                & \textbf{81.07}                 & \textbf{54.70}       & \textbf{86.31}       \\ \bottomrule
\end{tabular}
}
\caption{\small{\textbf{Main Results.} OOD detection performance comparison between \texttt{GradNorm} and baselines. All methods utilize the standard ResNetv2-101 model trained on ImageNet~\cite{deng2009imagenet}. The classification model is trained on {ID data only}.
$\uparrow$ indicates larger values are better, while $\downarrow$ indicates smaller values are better. All values are percentages. 
All methods are post hoc and can be directly used for pre-trained models.}}
\label{table:main}
\vspace{-0.4cm}
\end{table*}

\vspace{-0.2cm}
\subsection{Results and Ablation Studies}
\label{sec:ablations}
\vspace{-0.2cm}
\textbf{Comparison with output- and feature-based methods} 
The results for ImageNet evaluations are shown in Table~\ref{table:main}, where \texttt{GradNorm} demonstrates  superior performance. We report OOD detection performance for each OOD test dataset, as well as the average over the four datasets. For a fair comparison, all the methods use the same pre-trained backbone, without regularizing with auxiliary outlier data. 
In particular, we compare with MSP~\cite{hendrycks2016baseline}, ODIN~\cite{liang2018enhancing}, Mahalanobis~\cite{lee2018simple}, as well as Energy~\cite{liu2020energy}.
\emph{Details and hyperparameters of baseline methods can be found in Appendix~\ref{app:baseline}}. 

\texttt{GradNorm} outperforms the best output-based baseline, Energy score~\cite{liu2020energy}, by \textbf{16.33}\% in FPR95. \texttt{GradNorm} also outperforms a competitive feature-based method, Mahalanobis~\cite{lee2018simple}, by \textbf{26.99}\% in FPR95. We hypothesize that the increased size of label space makes the class-conditional Gaussian density estimation less viable. It is also worth noting that significant overheads can be introduced by some methods. 
For instance, Mahalanobis~\cite{lee2018simple} requires collecting feature representations from intermediate layers over the entire training set, which is expensive for large-scale datasets such as ImageNet. 
In contrast, \texttt{GradNorm} can be conveniently used through a simple gradient calculation without hyper-parameter tuning or additional training.
\paragraph{Gradients from the last layer is sufficiently informative} 
In this ablation, we investigate several variants of \texttt{GradNorm} where the gradients are extracted from different network depths. Specifically, we consider gradients of (1) \textbf{block n}: all trainable parameters in the $n$-th block, (2) \textbf{all parameters}: all trainable parameters from all layers of the network, 
and (3) \textbf{last layer parameters}: weight parameters from the last fully connected (FC) layer.

\begin{wraptable}{r}{0.4\textwidth}
{\footnotesize{
\begin{tabular}{c|cc}
\toprule
\multirow{2}{*}{\textbf{\begin{tabular}[c]{@{}c@{}}Gradient\\ Space\end{tabular}}} & \textbf{FPR95}       & \textbf{AUROC}     \\
                                                                                       & \multicolumn{1}{c}{$\downarrow$} & \multicolumn{1}{c}{$\uparrow$}  \\ \midrule
Block 1                                                                                & 73.52                & 76.41                \\
Block 2                                                                                & 74.34                & 76.63                \\
Block 3                                                                                & 71.73                & 78.11                \\
Block 4                                                                                & 65.07                & 85.11                \\
All params                                                                              & 69.35                & 81.14                \\
Last layer params                                                                             & \textbf{54.70}       & \textbf{86.31}       \\ \bottomrule
\end{tabular}}
}
    \caption{\small{Effect of \texttt{GradNorm} using different subset of gradients. Gradient norm derived from deeper layers yield better OOD detection performance.}}
    \label{tab:block_trend_ablation}
\end{wraptable}

Table~\ref{tab:block_trend_ablation} contrasts the OOD detection performance using different \emph{gradient space}. For each setting, we report the FPR95 and AUROC averaged across four OOD datasets. We observe that gradients from deeper layers tend to yield significantly better performance than shallower layers. This is desirable since gradients \wrt deeper layers are computationally more efficient than shallower layers. Interestingly,\texttt{GradNorm} obtained from the last linear layer yield the best results among all variants. Practically, one only needs to perform backpropagation \wrt the last linear layer, which incurs negligible computations. Therefore, our main results are based on the norm of gradients extracted from weight parameters in the last FC layer of the neural network.

\vspace{-0.2cm}
\paragraph{{GradNorm} with one-hot v.s. uniform targets}

In this ablation, we contrast \texttt{GradNorm} derived using uniform targets (ours) v.s. one-hot targets. Specifically, our scoring function Equation~\ref{eq:score}  is equivalent to 
\begin{align}
    S(\*x) = \lVert \frac{1}{C}\sum_{i=1}^C \frac{\partial \mathcal{L}_\text{CE}(f(\*x),i)}{\partial \*w} \rVert,
\end{align}
which captures the gradient of cross-entropy loss across \emph{all} labels. In contrast, we compare against an alternative scoring function that utilizes \emph{only one} dominant class label:
\begin{align}
    S_\text{one-hot}(\*x) = \lVert \frac{\partial \mathcal{L}_\text{CE}(f(\*x),\hat y)}{\partial \*w} \rVert,
\end{align}
where $\hat y$ is the predicted class with the largest output. 


\begin{figure*}[t]
    \centering
        \begin{subfigure}[b]{1.0\textwidth}
        \centering
        \includegraphics[width=1.0\textwidth]{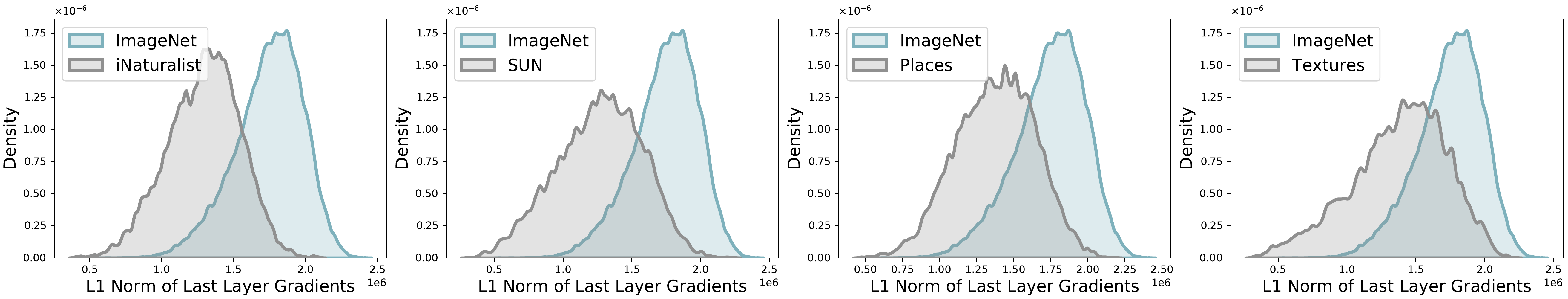}
        \caption{Gradient norms using KL divergence between the softmax prediction and the \textbf{uniform} target.}
        \label{subfig:score_dist_uniform}
        \end{subfigure}
        \begin{subfigure}[b]{1.0\textwidth}
        \centering
        \includegraphics[width=1.0\textwidth]{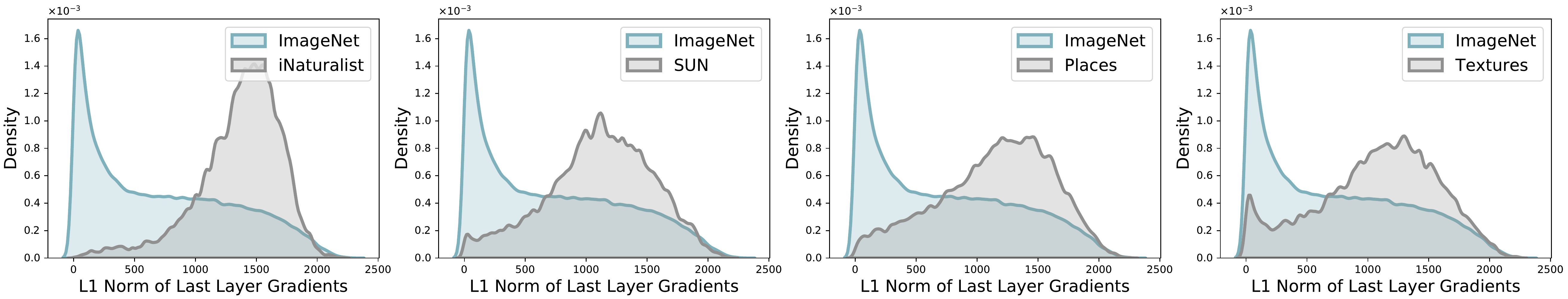}
        \caption{Gradient norms using KL divergence between the softmax prediction and the \textbf{one-hot} target.}
        \label{subfig:score_dist_onehot}
        \end{subfigure}
    \caption{\small{Comparison of $L_1$-norm distributions of last layer gradients between KL divergence with \emph{uniform} target and KL divergence with \emph{one-hot} target. We show in-distribution data in green and OOD data in gray. }}
    \label{fig:score_dist}
    \vspace{-0.4cm}
\end{figure*}

We first analyze the score distributions using uniform targets (\textbf{top}) and one-hot targets (\textbf{bottom}) for ID and OOD data in Figure~\ref{fig:score_dist}. 
There are two salient observations we can draw: (1) using uniform target (ours), gradients of ID data indeed have larger magnitudes than those of OOD data, as the softmax prediction tends to be less uniformly distributed (and therefore results in a larger KL divergence). In contrast, the gradient norm using one-hot targets shows the opposite trend, with ID data having lower magnitudes. This is also expected since the training objective explicitly minimizes the cross-entropy loss, which results in smaller gradients for the majority of ID data. (2) The score distribution using one-hot targets displays a strong overlapping between ID (green) and OOD (gray) data, with large variances. In contrast, our method \texttt{GradNorm} can significantly improve the separability between ID and OOD data, resulting in better OOD detection performance.

\begin{wrapfigure}{r}{0.48\textwidth}
\vspace{-0.6cm}
    \includegraphics[width=0.45\textwidth]{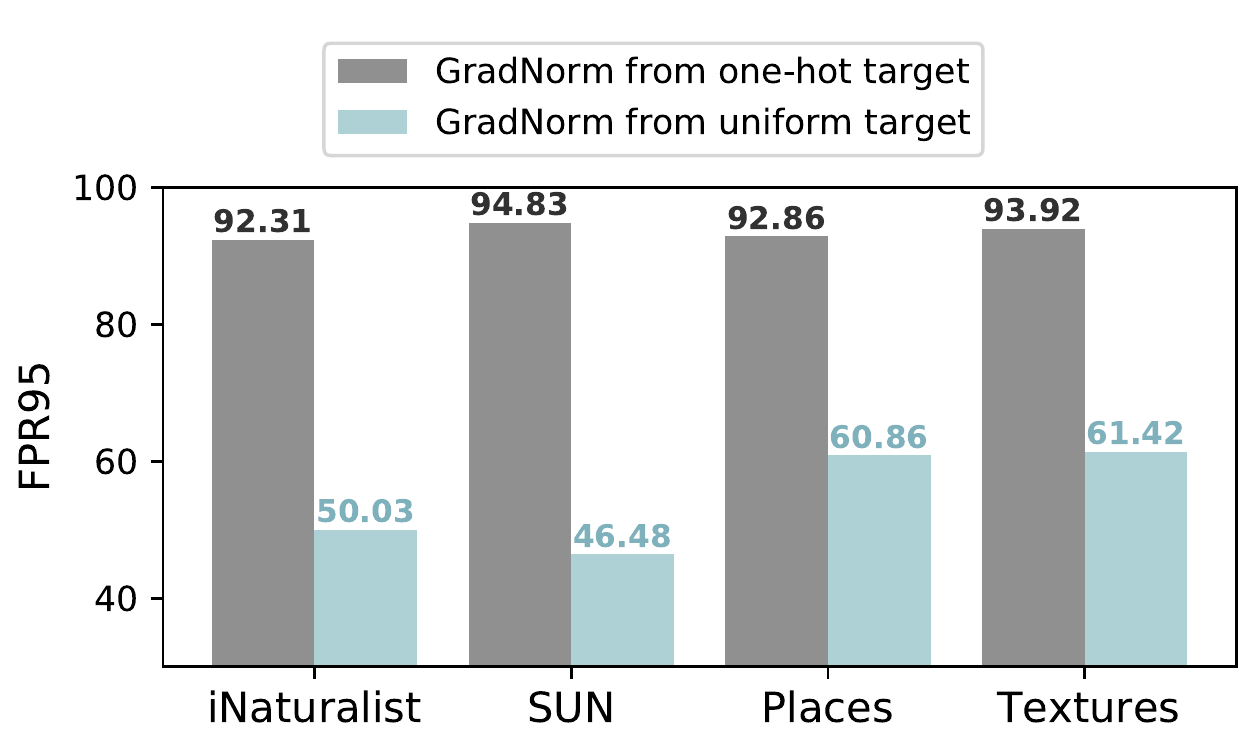}
    \caption{\small{OOD detection performance (FPR95) comparison between uniform (ours) v.s. one-hot target.}}
    \vspace{-0.4cm}
    \label{fig:traditional_ce_fpr95}
\end{wrapfigure}

Figure~\ref{fig:traditional_ce_fpr95} reports the OOD detection performance using uniform targets (ours) v.s. one-hot targets. We use $L_1$-norm in both cases. For one-hot targets, we use the negative norm, \ie $-S_\text{one-hot}(\*x)$, to align with the convention that ID data has higher scores. \texttt{GradNorm} with uniform targets outperforms its counterpart with one-hot targets by a large margin. For instance, \texttt{GradNorm} reduces FPR95 by \textbf{48.35}\% when evaluated on the SUN dataset. Our analysis signifies the importance of measuring OOD uncertainty using all label information.


\paragraph{GradNorm is effective on alternative neural network architecture} We evaluate \texttt{GradNorm} on a different architecture DenseNet-121~\cite{huang2017densely}, and report performance in Table~\ref{tab:densenet}. \texttt{GradNorm} is consistently effective, outperforming the best baseline, Energy~\cite{liu2020energy}, by \textbf{10.29}\% in FPR95.

\begin{table}[h]
    \centering
\scriptsize{
\begin{tabular}{c|l|C{0.03\textwidth}C{0.045\textwidth}|C{0.03\textwidth}C{0.045\textwidth}|C{0.03\textwidth}C{0.045\textwidth}|C{0.03\textwidth}C{0.045\textwidth}|C{0.03\textwidth}C{0.045\textwidth}}
\toprule
\multirow{3}{*}{\textbf{\begin{tabular}[c]{@{}c@{}}Method\\ Space\end{tabular}}} & \multicolumn{1}{c|}{\multirow{3}{*}{\textbf{Method}}} & \multicolumn{2}{c|}{\textbf{iNaturalist}}    & \multicolumn{2}{c|}{\textbf{SUN}}            & \multicolumn{2}{c|}{\textbf{Places}}         & \multicolumn{2}{c|}{\textbf{Textures}}       & \multicolumn{2}{c}{\textbf{Average}}        \\ \cline{3-12} 
                                                                                 & \multicolumn{1}{c|}{}                                 & \tiny{FPR95}                & \tiny{AUROC}                 & \tiny{FPR95}                & \tiny{AUROC}              & \tiny{FPR95}                & \tiny{AUROC}                 & \tiny{FPR95}                & \tiny{AUROC}                 & \tiny{FPR95}                & \tiny{AUROC}               \\
                                                                                      & \multicolumn{1}{c|}{}                                 & \multicolumn{1}{c}{$\downarrow$} & \multicolumn{1}{c|}{$\uparrow$} & \multicolumn{1}{c}{$\downarrow$} & \multicolumn{1}{c|}{$\uparrow$} & \multicolumn{1}{c}{$\downarrow$} & \multicolumn{1}{c|}{$\uparrow$} & \multicolumn{1}{c}{$\downarrow$} & \multicolumn{1}{c|}{$\uparrow$} & \multicolumn{1}{c}{$\downarrow$} & \multicolumn{1}{c}{$\uparrow$}  \\ \midrule
\multirow{3}{*}{Output}                                                          & MSP\tiny{~\cite{hendrycks2016baseline}}                                                   & 48.55                & 89.16                 & 69.39                & 80.46                 & 71.42                & 80.11                 & 68.51                & 78.69                 & 64.47                & 82.11                \\
                                                                                 & ODIN\tiny{~\cite{liang2018enhancing}}                                                  & 37.00                & 93.29                 & 57.30                & 86.12                 & 61.91                & 84.14                 & 56.49                & 84.62                 & 53.18                & 87.04                \\
                                                                                 & Energy\tiny{~\cite{liu2020energy}}                                                & 36.39                & 93.29                 & 54.91                & 86.53                 & 59.98                & \textbf{84.29}        & 53.87                & 85.07                 & 51.29                & 87.30                \\ \midrule
Feature                                                                          & Mahalanobis\tiny{~\cite{lee2018simple}}                                           &  97.36 &	42.24 &	98.24 &	41.17 &	97.32 &	47.27 &	62.78 &	56.53 &	88.93 &	46.80                 \\ \midrule
 \multirow{1}{*}{Gradient}                                                      
                                                                                 & \textbf{GradNorm (ours)}                              & \textbf{23.87}       & \textbf{93.97}        & \textbf{43.04}       & \textbf{87.79}        & \textbf{53.92}       & 83.04                 & \textbf{43.16}       & \textbf{87.48}        & \textbf{41.00}       & \textbf{88.07}       \\ \bottomrule
\end{tabular}
}
    \caption{\small OOD detection performance comparison on a different architecture, \textbf{DenseNet-121}~\cite{huang2017densely}. Model is trained on ImageNet-1k~\cite{deng2009imagenet} as the ID dataset. 
    All methods are post hoc and can be directly used for pre-trained models.}
    \label{tab:densenet}
\end{table}



\begin{figure}[h]
    \centering
    \includegraphics[width=0.99\textwidth]{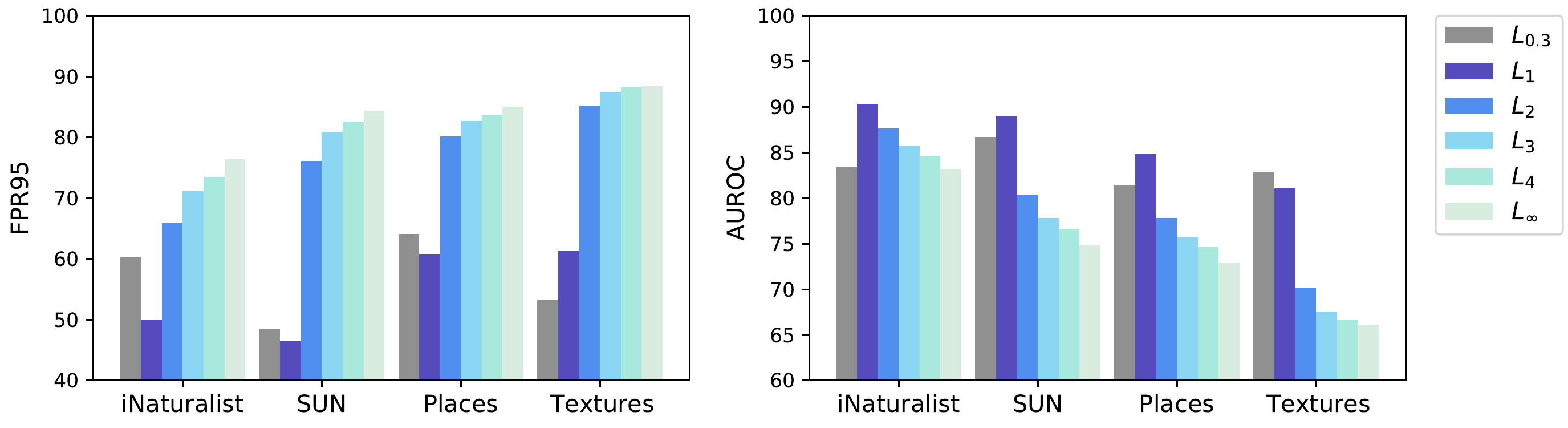}
    \caption{\small{OOD detection performance comparison under different $L_p$-norms. We show FPR95 (\emph{left}) and AUROC (\emph{right}).}}
    \label{fig:norm_ablation}
\end{figure}

\vspace{-0.2cm}
\paragraph{$L_1$-norm is the most effective} How does the choice of $L_p$-norm in Equation~\ref{eq:score} affect the OOD detection performance? To understand this, we show in Figure~\ref{fig:norm_ablation} the comparison using $L_{1 \sim 4}$-norm, $L_{\infty}$-norm, as well as the fraction norm (with $p=0.3$). 
Compared with higher-order norms, $L_1$-norm achieves the best OOD detection performance on all four datasets. We hypothesize that $L_1$-norm is better suited since it captures information equally from all dimensions in the gradient space, whereas higher-order norms will unfairly highlight larger elements rather than smaller elements (due to the effect of the exponent $p$). In the extreme case, $L_{\infty}$-norm only considers the largest element (in absolute value) and results in the worst OOD detection performance among all norms. 
On the other hand, the fraction norm overall does not outperform $L_1$-norm. \emph{We additionally provide results for more $L_p$-norms in Appendix~\ref{app:more_norm}, with $p=\{0.3, 0.5, 0.8, 1,2,3,4,5,6,\infty\}$}.

\begin{figure}[h]
\centering
    \vspace{-0.2cm}
    \includegraphics[width=0.95\textwidth]{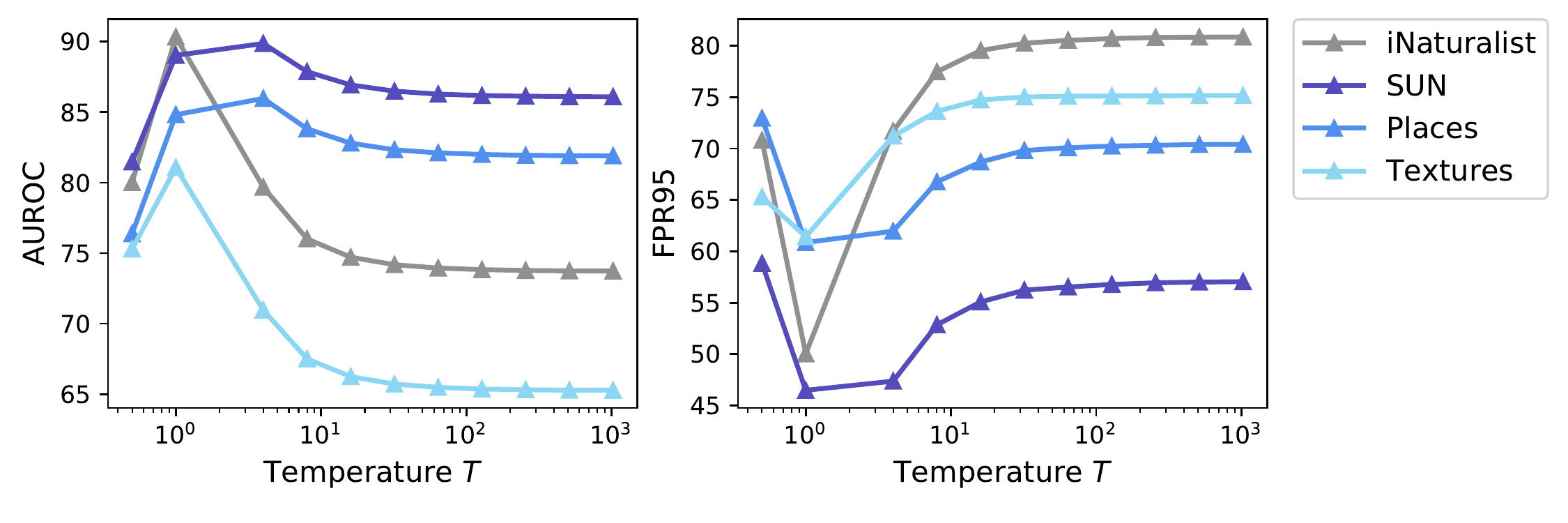}
    \caption{\small{OOD detection performance of \texttt{GradNorm} with varying temperature parameter $T$. We show AUROC (\emph{left}) and FPR95 (\emph{right}).}}
    \label{fig:temper_ablation}
\end{figure}

\textbf{Effect of temperature scaling} We evaluate our method \texttt{GradNorm} with different temperatures $T$ from $T=0.5$ to $T=1024$. As shown in Figure~\ref{fig:temper_ablation}, $T=1$ is optimal, while either increasing or decreasing the temperature will degrade the performance. This can be explained mathematically via the $V$ term in Equation~\ref{eq:decomp}. Specifically, using a large temperature will result in a smoother softmax distribution, with $C \cdot \frac{e^{f_j / T}}{\sum_{j=1}^C e^{{f_{j}} / T}}$ closer to 1 (and $V\rightarrow 0$). This leads to a less distinguishable distribution between ID and OOD. Our method can be hyperparameter-free by setting $T=1$. For completeness, we have included numerical results under a wider range of $T$ in Appendix~\ref{app:more_temper}.

\begin{wrapfigure}{r}{0.45\textwidth}
\vspace{-1.5cm}
    \includegraphics[width=0.43\textwidth]{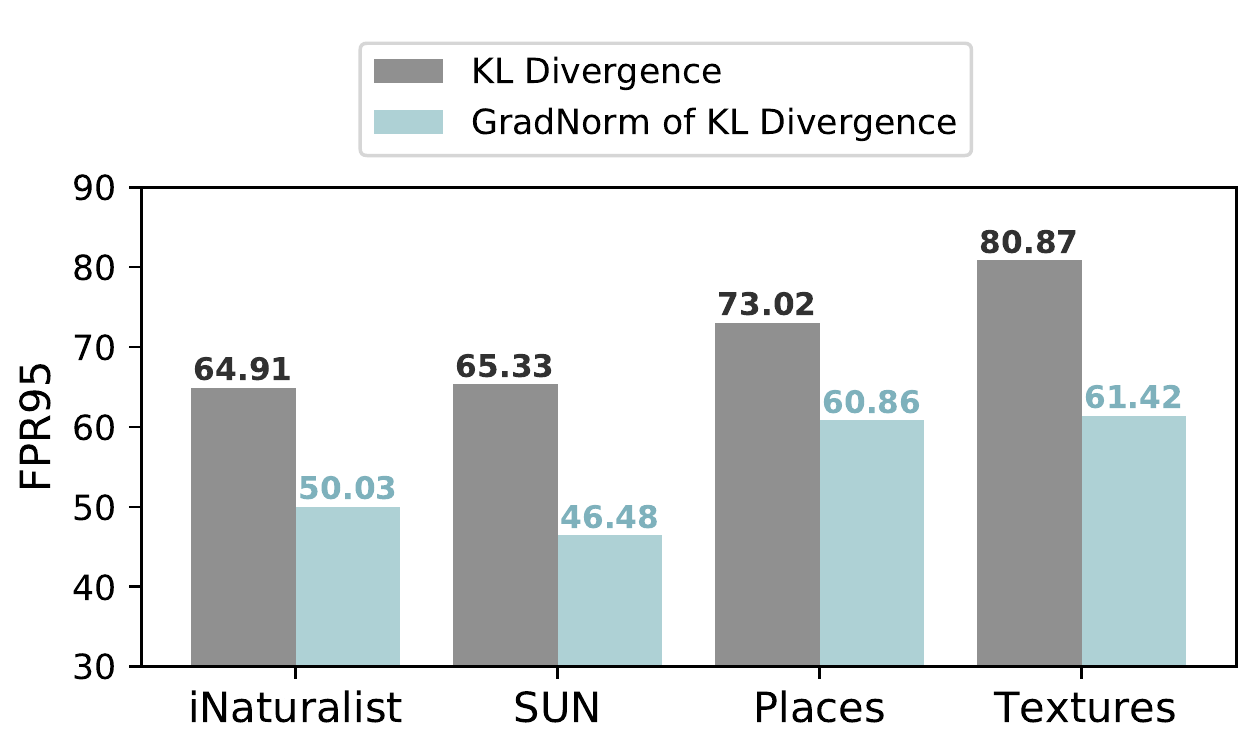}
    \vspace{-0.3cm}
    \caption{\small{Comparison between \texttt{GradNorm} v.s. directly using the KL divergence as scoring function.}}
    \vspace{-0.5cm}
    \label{fig:kl_ablation}
\end{wrapfigure}

\vspace{-0.2cm}
\paragraph{{GradNorm} is more effective than directly using KL divergence} We provide an ablation, contrasting the performance of using \texttt{GradNorm} v.s. using the KL divergence derived from Equation~\ref{eq:kl} as OOD scoring function. The results are show in Figure~\ref{fig:kl_ablation}, where \texttt{GradNorm} yields significantly better performance than the KL divergence directly extracted from the \emph{output space}, demonstrating the superiority of \emph{gradient space} for OOD detection.

\begin{wraptable}{r}{0.4\textwidth}
\vspace{-0.3cm}
    \centering
\footnotesize{
\begin{tabular}{c|cc}
\toprule
\multirow{2}{*}{\textbf{\begin{tabular}[c]{@{}c@{}}Model Size\\ (depth x width)\end{tabular}}} & \textbf{FPR95}          & \textbf{AUROC}            \\
                                                                                                      & \multicolumn{1}{c}{$\downarrow$}     & \multicolumn{1}{c}{$\uparrow$}\\ \midrule
50x1                                                                                                  & 56.91          & 84.17  \\
101x1                                                                                                 & \textbf{55.84} & \textbf{84.63} \\
50x3                                                                                                  & 61.74          & 81.89 \\
152x2                                                                                                 & 61.76          & 81.33  \\
101x3                                                                                                 & 66.20          & 78.89 \\ \bottomrule
\end{tabular}
}
\caption{\small{OOD detection performance as the model capacity increases.}}
\label{tab:arch_ablation}
\vspace{-0.2cm}
\end{wraptable}

\paragraph{Effect of model capacity} In this ablation, we explore the OOD detection performance of \texttt{GradNorm} with varying model capacities. For the ease of experiments, we directly use Google BiT-S models pre-trained on ImageNet-1k~\cite{deng2009imagenet}. We compare the performance of the following model family (in increasing size): BiT-S-R50x1, BiT-S-R101x1, BiT-S-R50x3, BiT-S-R152x2, BiT-S-R101x3. All models are ResNetv2 architectures with varying depths and width factors. The average performance on 4 OOD datasets is reported in Table~\ref{tab:arch_ablation}. OOD detection performance is optimal when the model size is relatively small (ResNetv2-101x1), while further increasing model capacity will degrade the performance. Our experiments suggest that overparameterization can make gradients less distinguishable between ID and OOD data and that \texttt{GradNorm} is more suitable under a mild model capacity.

\section{Analysis of Gradient-based Method}
\label{sec:analysis}
In this section, we analyze the best variant of \texttt{GradNorm}, $L_1$-norm of the last layer gradients (see Section~\ref{sec:ablations}), and provide insights on the mathematical interpretations.
Specifically, we denote the last FC layer in a neural network by: 
\begin{equation}
    f(\*x) = \*W^\top\*x + \mathbf{b},
\end{equation}
where $f = [f_1, f_2, \dots, f_C]^\top \in \mathbb{R}^C$ is the logit output, $\mathbf{x} = [x_1, x_2, \dots, x_m]^\top \in \mathbb{R}^m$ is the input feature vector, $\mathbf{W} \in \mathbb{R}^{m \times C}$ is the weight matrix, and $\mathbf{b} \in \mathbb{R}^C$ is the bias vector.


\begin{figure*}[t]
    \centering
    \begin{subfigure}{1\textwidth}
    \includegraphics[width=1.0\textwidth]{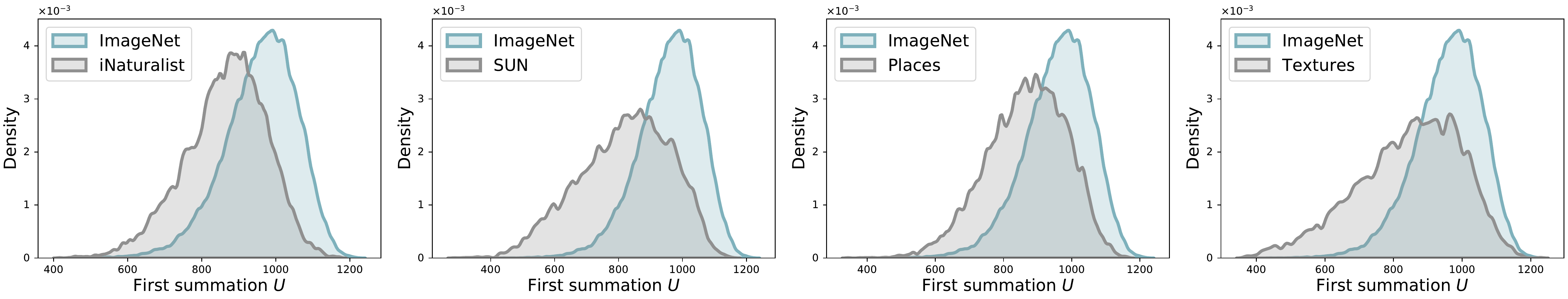}
    \caption{Distribution of  $U$}
    \end{subfigure}
    \begin{subfigure}{1\textwidth}
    \includegraphics[width=1.0\textwidth]{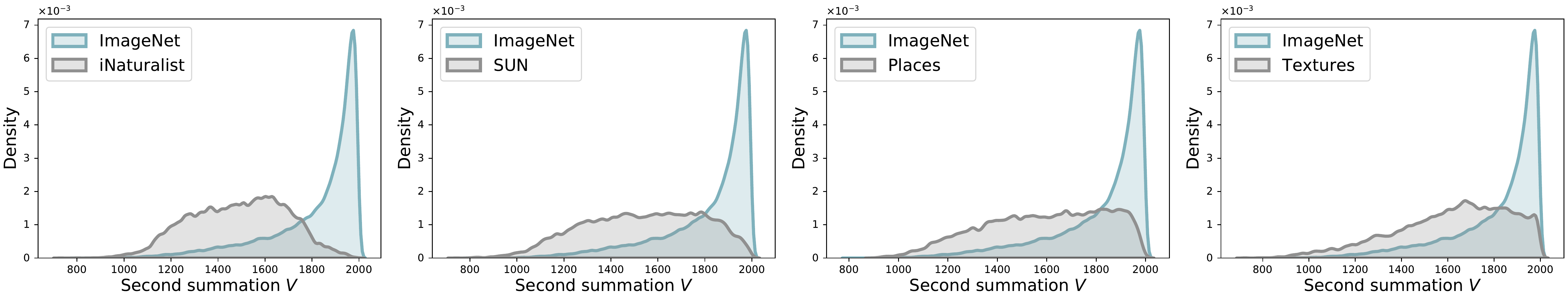}
    \caption{Distribution of $V$}
    \end{subfigure}
    \caption{\small{We show the distributions of the two summations decomposed from the $L_1$-norm of the last layer gradient, for both in-distribution data (blue) and out-of-distribution data (gray). }}
    \label{fig:first_and_second_sum}
\end{figure*}

\textbf{{GradNorm} captures joint information between feature and output} First we can rewrite the KL divergence between the softmax prediction and the uniform target as:
\begin{align*}
        D_\text{KL} (\*u \lVert \text{softmax}(f(\*x))) & = - \frac{1}{C}\sum_{c=1}^C \log{ \frac{e^{{f_c} / T}}{\sum_{j=1}^C e^{{f_{j}} / T}}} - H(\*u)  \\&= - \frac{1}{C} \left( \frac{1}{T}\sum_{c=1}^C f_c - C \cdot \log{{\sum_{j=1}^C e^{{f_{j}} / T}}} \right) - H(\*u).
\end{align*}

Then we consider the derivative of $D_\text{KL}$ \wrt each output logit $f_c$:
\begin{align*}
        \frac{\partial D_\text{KL}}{\partial f_c}
       & = -\frac{1}{CT} \left( 1 - CT \cdot \frac{\partial \left( \log{{\sum_{j=1}^C e^{{f_{j}} / T}}}\right)}{\partial f_c} \right) \\
    &    = -\frac{1}{CT} \left( 1 - C \cdot \frac{e^{f_c / T}}{\sum_{j=1}^C e^{{f_{j}} / T}} \right).
\end{align*}

Next the derivative of $ D_\text{KL}$ \wrt the weight matrix can be written as:
\begin{equation*}
        \frac{\partial  D_\text{KL}}{\partial \mathbf{W}} = \mathbf{x} \frac{\partial  D_\text{KL}}{\partial f} 
        = -\frac{1}{CT} \cdot [x_1, x_2, \dots, x_m]^{\top}
        [1 - C \cdot \frac{e^{f_1 / T}}{\sum_{j=1}^C e^{{f_{j}} / T}}, \dots, 1 - C \cdot \frac{e^{f_C / T}}{\sum_{j=1}^C e^{{f_{j}} / T}}].
\end{equation*}

Finally, the $L_1$-norm of gradients of the weight matrix is simply the sum of absolute values of all elements in the gradient matrix:
\begin{equation}
    \begin{split}
        S(\*x) &= \sum_{i=1}^m \sum_{j=1}^C \left|\left(\frac{\partial  D_\text{KL}}{\partial \mathbf{W}}\right)_{ij}\right| = \frac{1}{CT} \sum_{i=1}^m \left( |x_i| \left(\sum_{j=1}^C \left|1 - C \cdot \frac{e^{f_j / T}}{\sum_{j=1}^C e^{{f_{j}} / T}}\right|\right)\right) \\
        &= \frac{1}{CT}  \left(\sum_{i=1}^m |x_i|\right) \left(\sum_{j=1}^C \left|1 - C \cdot \frac{e^{f_j / T}}{\sum_{j=1}^C e^{{f_{j}} / T}}\right|\right) \\
        & \triangleq \frac{1}{CT} U \cdot V,
    \end{split}
    \label{eq:decomp}
\end{equation}
where the first multiplicative term $U=\sum_{i=1}^m |x_i|$ is the $L_1$-norm of the feature vector $\*x$, and the second term $V$ characterizes information in the output space.

\textbf{Ablation on $U$ and $V$} In Figure~\ref{fig:first_and_second_sum} we plot distribution densities of $U$ and $V$, for both ID and OOD data. 
It is important to note that $U$ and $V$ measure statistical distributions in the feature space and the output space, respectively. Therefore, \texttt{GradNorm} captures the joint information between the feature and the output space. The multiplication of both $U$ and $V$ results in an overall stronger separability between ID and OOD, as seen in Figure~\ref{subfig:score_dist_uniform}. We report the OOD detection performance using $U$ and $V$ individually as scoring functions in Table~\ref{tab:uv_ood_performance}, both of which are less competitive than \texttt{GradNorm}.

\begin{table}[h]
    \centering
\scriptsize{
\begin{tabular}{c|C{0.04\textwidth}C{0.05\textwidth}|C{0.04\textwidth}C{0.05\textwidth}|C{0.04\textwidth}C{0.05\textwidth}|C{0.04\textwidth}C{0.05\textwidth}|C{0.04\textwidth}C{0.05\textwidth}}
\toprule
 \multicolumn{1}{c|}{\multirow{3}{*}{\textbf{Method}}} & \multicolumn{2}{c|}{\textbf{iNaturalist}}    & \multicolumn{2}{c|}{\textbf{SUN}}            & \multicolumn{2}{c|}{\textbf{Places}}         & \multicolumn{2}{c|}{\textbf{Textures}}       & \multicolumn{2}{c}{\textbf{Average}}        \\ 
\cline{2-11} 
                                                                                  \multicolumn{1}{c|}{}                                 & \scriptsize{FPR95}                & \scriptsize{AUROC}                 & \scriptsize{FPR95}                & \scriptsize{AUROC}              & \tiny{FPR95}                & \scriptsize{AUROC}                 & \scriptsize{FPR95}                & \scriptsize{AUROC}                 & \scriptsize{FPR95}                & \scriptsize{AUROC}               \\
                                                                                      \multicolumn{1}{c|}{}                                 & \multicolumn{1}{c}{$\downarrow$} & \multicolumn{1}{c|}{$\uparrow$} & \multicolumn{1}{c}{$\downarrow$} & \multicolumn{1}{c|}{$\uparrow$} & \multicolumn{1}{c}{$\downarrow$} & \multicolumn{1}{c|}{$\uparrow$} & \multicolumn{1}{c}{$\downarrow$} & \multicolumn{1}{c|}{$\uparrow$} & \multicolumn{1}{c}{$\downarrow$} & \multicolumn{1}{c}{$\uparrow$}  \\ \midrule
                                                                                    U (feature space) &  77.84 & 74.33 & 61.90 & 78.74 & 76.42 & 72.75 & 67.84 & 72.77 & 71.00 & 74.65\\
                                                                                    V (output space) & 66.14 & 88.45 & 69.49 & 83.13 & 75.95 & 78.98 & 81.13 & 76.06 & 73.18 & 81.66  \\
                                                                                   \textbf{ U $\cdot$ V (Joint space)} & \textbf{50.05} & \textbf{90.33} & \textbf{46.48} & \textbf{89.03} & \textbf{60.86} & \textbf{84.82} & \textbf{61.42} & \textbf{81.07} & \textbf{54.70} & \textbf{86.31} \\
                                                            
\bottomrule
\end{tabular}
}
    \caption{\small{OOD detection performance using the decomposed $U$ (feature space) and $V$ (output space) as scoring functions. Model is ResNetv2-101 trained on ImageNet-1k~\cite{deng2009imagenet}.}}
    \label{tab:uv_ood_performance}
    \vspace{-0.5cm}
\end{table}

\vspace{-0.2cm}
\section{Discussion}
\label{sec:discussion}
\vspace{-0.2cm}
To the best of our knowledge, there is very limited prior work studying how to use gradients for OOD detection.  In this section, we discuss connections and differences between \texttt{GradNorm} and previous OOD detection approaches that utilize gradient information, in particular ODIN (Section~\ref{sec:connection_with_odin}) and \citeauthor{lee2020gradients}'s approach (Section~\ref{sec:connection_with_al}).


\vspace{-0.2cm}
\subsection{Comparison with ODIN}
\label{sec:connection_with_odin}
\vspace{-0.1cm}

Our work is inspired by ODIN~\cite{liang2018enhancing}, which first explored using gradient information for OOD detection. In particular, ODIN proposed using input pre-processing by adding small perturbations obtained from the input gradients. The goal of ODIN perturbations is to increase the softmax score of any given input by reinforcing the model's belief in the predicted label. Ultimately the perturbations have been found to create a greater gap between the softmax scores of ID and OOD inputs, thus making them more separable and improving the performance of OOD detection.

It is important to note that ODIN only uses gradients \emph{implicitly} through input perturbation, and OOD scores are still derived from the output space of the perturbed inputs. Different from ODIN, \texttt{GradNorm} utilizes information solely obtained from the {gradient space}. The effectiveness of \texttt{GradNorm} beckons a revisiting of combining information obtainable from the gradient space and the output space, which could provide a stronger method. We leave this question for future exploration.

\vspace{-0.2cm}
\subsection{Comparison with \citeauthor{lee2020gradients}}
\label{sec:connection_with_al}
\vspace{-0.1cm}
\citeauthor{lee2020gradients}~\cite{lee2020gradients} proposed to train an auxiliary binary classifier using gradient information from ID and OOD data. Importantly, they do not directly use gradient norms for OOD detection, but instead, use them as the input for training a separate binary classifier. Furthermore, the binary classifier is trained on the OOD datasets, which can unfairly overfit the test data and does not suit OOD-agnostic settings in the real world. In contrast, our methodology mitigates the shortcomings in that \texttt{GradNorm} (1) does not require any new model training, (2) is hyperparameter-free, and (3) is suitable for OOD-agnostic settings. For these reasons, these two methods are not directly comparable. However, for completeness, we also reproduce \citeauthor{lee2020gradients}'s method using random noise as a surrogate of OOD data and compare it with \texttt{GradNorm} in Appendix~\ref{app:comparison_with_la}. \texttt{GradNorm} outperforms their approach by 15.45\% in FPR95 in this fair comparison.

Moreover, we provide comprehensive ablation studies and analyses on different design choices in using gradient-based methods for OOD detection (network architectures, gradients at different layers, loss functions for backpropagation, different $L_p$-norms, diverse evaluation datasets, and different temperatures, etc.), which were previously not studied in~\cite{lee2020gradients}. In particular, \citeauthor{lee2020gradients} utilize $L_2$-norm gradients without comparing them with other norms. Our ablation study leads to the new finding that $L_1$-norm works best among all variants with \texttt{GradNorm}, and outperforms $L_2$-norm by up to 22.31\% in FPR95. Furthermore, \citeauthor{lee2020gradients} utilize gradients from all layers to train a separate binary classifier, which can cause the computational cost to become intractable for deeper and larger models. In contrast, with \texttt{GradNorm} we show that the last layer gradient will always yield the best performance among all gradient set selections. Consequently, \texttt{GradNorm} incurs negligible computational cost. We believe such thorough understandings will be valuable for the field.

\vspace{-0.1cm}
\section{Related Work}
\label{sec:related_work}
\vspace{-0.3cm}

\paragraph{OOD uncertainty estimation with discriminative models}  The problem of classification with rejection can date back to early works on abstention~\cite{chow1970optimum, fumera2002support}, which considered simple model families such as SVMs~\cite{cortes1995support}.
The phenomenon of neural networks' overconfidence in out-of-distribution data is first revealed by \citeauthor{nguyen2015deep}~\cite{nguyen2015deep}.
Early works attempted to improve the OOD uncertainty estimation by proposing the ODIN score~\citep{hsu2020generalized, liang2018enhancing} and Mahalanobis
distance-based confidence score~\citep{lee2018simple}.
Recent work by \citeauthor{liu2020energy}~\citep{liu2020energy} proposed using an energy score for OOD uncertainty estimation, which can be easily derived from a discriminative classifier and demonstrated advantages over the softmax confidence score both empirically and theoretically. ~\citeauthor{wang2021canmulti}~\cite{wang2021canmulti} further showed an energy-based approach  can improve OOD uncertainty estimation for multi-label classification networks. \citeauthor{huang2021mos}~\cite{huang2021mos} revealed that approaches developed for common CIFAR benchmarks might not translate effectively into a large-scale ImageNet benchmark, highlighting the need to evaluate OOD uncertainty estimation in a large-scale real-world setting.  
Existing approaches derive OOD scores from either output or feature space. In contrast, we show that \emph{gradient space} carries surprisingly useful information for OOD uncertainty estimation, which was underexplored in the literature.

\vspace{-0.2cm}
\paragraph{OOD uncertainty estimation with generative models} ~ Alternative approaches for detecting OOD inputs resort to generative models that directly estimate density~\cite{dinh2017density,huang2017stacked, kingma2014autoencoding, oord2016conditional, rezende2014stochastic,  tabak2013family}. An input is deemed as OOD if it lies in the low-likelihood regions. A plethora of literature has emerged to utilize generative models for OOD detection~\cite{ kirichenko2020normalizing, ren2019likelihood, schirrmeister2020understanding, serra2019input, wang2020further, winkens2020contrastive, xiao2020likelihood}.
Interestingly, \citeauthor{nalisnick2018deep}~\cite{nalisnick2018deep} showed that deep generative models can assign a high likelihood to OOD data.
Moreover, generative models can be prohibitively challenging to train and optimize, and the performance can often lag behind the discriminative counterpart. In contrast, our method relies on a discriminative classifier, which is easier to optimize and achieves stronger performance. 



\vspace{-0.2cm}
\paragraph{Distributional shifts} Distributional shifts have attracted increasing research interests. It is important to recognize and differentiate various types of distributional shift problems. Literature in OOD detection is commonly concerned about model reliability and detection of label-space shifts~\cite{hendrycks2016baseline, liang2018enhancing, liu2020energy}, where the OOD inputs have disjoint labels \wrt ID data and therefore \emph{should not be predicted by the model}. Meanwhile, some works considered covariate shifts in the input space~\citep{hendrycks2019benchmarking, malinin2021shifts, ovadia2019can}, where inputs can be corruption-shifted or domain-shifted~\citep{hsu2020generalized}. However, covariate shifts are commonly used to evaluate model robustness and domain generalization performance, where the label space $\mathcal{Y}$ remains the same during test time. It is important to note that our work focuses on the detection of shifts where the model should not make any prediction, instead of covariate shifts where the model is expected to \emph{generalize}. 


\vspace{-0.3cm}
\section{Conclusion}
\vspace{-0.2cm}
\label{sec:conclusion}
In this paper, we propose \texttt{GradNorm}, a novel OOD uncertainty estimation approach utilizing information extracted from the \emph{gradient space}. 
Experimental results show that our gradient-based method can improve the performance of OOD detection by up to  {16.33}\% in FPR95, establishing superior performance.
Extensive ablations provide further understandings of our approach. 
We hope that our research brings to light the informativeness of gradient space, and inspires future work to utilize gradient space for OOD uncertainty estimation.

\vspace{-0.2cm}
\section{Societal Impact}
\label{sec:broad}
\vspace{-0.3cm}
Our project aims to improve the reliability and safety of modern machine learning models.
This stands to benefit a wide range of fields and societal activities. We believe out-of-distribution uncertainty estimation is an increasingly critical component of systems that range from consumer and business applications (e.g., digital content understanding) to transportation (e.g., driver assistance systems and autonomous vehicles), and to health care (e.g., unseen disease identification). Many of these applications require classification models in operation.\@
Through this work and by releasing our code, we hope to provide machine learning researchers with a new methodological perspective and offer machine learning practitioners an easy-to-use tool that renders safety against OOD data in the real world. While we do not anticipate any negative consequences to our work, we hope to continue to build on our framework in future work. 
\vspace{-0.2cm}
\section*{Acknowledgement}
\vspace{-0.3cm}
 Research is supported by the Office of the Vice Chancellor for Research and Graduate Education (OVCRGE) with funding from the Wisconsin Alumni Research Foundation (WARF).


\newpage
\bibliography{neurips_2021}

\begin{thebibliography}{48}
\providecommand{\natexlab}[1]{#1}
\providecommand{\url}[1]{\texttt{#1}}
\expandafter\ifx\csname urlstyle\endcsname\relax
  \providecommand{\doi}[1]{doi: #1}\else
  \providecommand{\doi}{doi: \begingroup \urlstyle{rm}\Url}\fi

\bibitem[Chen et~al.(2021)Chen, Li, Wu, Liang, and Jha]{chen2021robustifying}
Jiefeng Chen, Yixuan Li, Xi~Wu, Yingyu Liang, and Somesh Jha.
\newblock Atom: Robustifying out-of-distribution detection using outlier
  mining.
\newblock \emph{In Proceedings of European Conference on Machine Learning and
  Principles and Practice of Knowledge Discovery in Databases (ECML PKDD)},
  2021.

\bibitem[Chow(1970)]{chow1970optimum}
CK~Chow.
\newblock On optimum recognition error and reject tradeoff.
\newblock \emph{IEEE Transactions on information theory}, 16\penalty0
  (1):\penalty0 41--46, 1970.

\bibitem[Cimpoi et~al.(2014{\natexlab{a}})Cimpoi, Maji, Kokkinos, Mohamed, ,
  and Vedaldi]{cimpoi14describing}
M.~Cimpoi, S.~Maji, I.~Kokkinos, S.~Mohamed, , and A.~Vedaldi.
\newblock Describing textures in the wild.
\newblock In \emph{Proceedings of the {IEEE} Conf. on Computer Vision and
  Pattern Recognition ({CVPR})}, 2014{\natexlab{a}}.

\bibitem[Cimpoi et~al.(2014{\natexlab{b}})Cimpoi, Maji, Kokkinos, Mohamed, and
  Vedaldi]{cimpoi2014describing}
Mircea Cimpoi, Subhransu Maji, Iasonas Kokkinos, Sammy Mohamed, and Andrea
  Vedaldi.
\newblock Describing textures in the wild.
\newblock In \emph{Proceedings of the IEEE Conference on Computer Vision and
  Pattern Recognition}, pages 3606--3613, 2014{\natexlab{b}}.

\bibitem[Cortes and Vapnik(1995)]{cortes1995support}
Corinna Cortes and Vladimir Vapnik.
\newblock Support-vector networks.
\newblock \emph{Machine learning}, 20\penalty0 (3):\penalty0 273--297, 1995.

\bibitem[Deng et~al.(2009)Deng, Dong, Socher, Li, Li, and
  Fei-Fei]{deng2009imagenet}
Jia Deng, Wei Dong, Richard Socher, Li-Jia Li, Kai Li, and Li~Fei-Fei.
\newblock Imagenet: A large-scale hierarchical image database.
\newblock In \emph{2009 IEEE conference on computer vision and pattern
  recognition}, pages 248--255. Ieee, 2009.

\bibitem[Dinh et~al.(2017)Dinh, Sohl{-}Dickstein, and Bengio]{dinh2017density}
Laurent Dinh, Jascha Sohl{-}Dickstein, and Samy Bengio.
\newblock Density estimation using real {NVP}.
\newblock In \emph{5th International Conference on Learning Representations,
  {ICLR} 2017}, 2017.

\bibitem[Fumera and Roli(2002)]{fumera2002support}
Giorgio Fumera and Fabio Roli.
\newblock Support vector machines with embedded reject option.
\newblock In \emph{International Workshop on Support Vector Machines}, pages
  68--82. Springer, 2002.

\bibitem[Goodfellow et~al.(2015)Goodfellow, Shlens, and
  Szegedy]{goodfellow2014explaining}
Ian~J. Goodfellow, Jonathon Shlens, and Christian Szegedy.
\newblock Explaining and harnessing adversarial examples.
\newblock In \emph{International Conference on Learning Representations, {ICLR}
  2015}, 2015.

\bibitem[He et~al.(2016{\natexlab{a}})He, Zhang, Ren, and Sun]{he2016deep}
Kaiming He, Xiangyu Zhang, Shaoqing Ren, and Jian Sun.
\newblock Deep residual learning for image recognition.
\newblock In \emph{Proceedings of the IEEE conference on computer vision and
  pattern recognition}, pages 770--778, 2016{\natexlab{a}}.

\bibitem[He et~al.(2016{\natexlab{b}})He, Zhang, Ren, and Sun]{he2016identity}
Kaiming He, Xiangyu Zhang, Shaoqing Ren, and Jian Sun.
\newblock Identity mappings in deep residual networks.
\newblock In \emph{European conference on computer vision}, pages 630--645.
  Springer, 2016{\natexlab{b}}.

\bibitem[Hendrycks and Dietterich(2019)]{hendrycks2019benchmarking}
Dan Hendrycks and Thomas Dietterich.
\newblock Benchmarking neural network robustness to common corruptions and
  perturbations.
\newblock \emph{arXiv preprint arXiv:1903.12261}, 2019.

\bibitem[Hendrycks and Gimpel(2017)]{hendrycks2016baseline}
Dan Hendrycks and Kevin Gimpel.
\newblock A baseline for detecting misclassified and out-of-distribution
  examples in neural networks.
\newblock In \emph{5th International Conference on Learning Representations,
  {ICLR} 2017}, 2017.

\bibitem[Hsu et~al.(2020)Hsu, Shen, Jin, and Kira]{hsu2020generalized}
Yen-Chang Hsu, Yilin Shen, Hongxia Jin, and Zsolt Kira.
\newblock Generalized odin: Detecting out-of-distribution image without
  learning from out-of-distribution data.
\newblock In \emph{Proceedings of the IEEE/CVF Conference on Computer Vision
  and Pattern Recognition}, pages 10951--10960, 2020.

\bibitem[Huang et~al.(2017{\natexlab{a}})Huang, Liu, Van Der~Maaten, and
  Weinberger]{huang2017densely}
Gao Huang, Zhuang Liu, Laurens Van Der~Maaten, and Kilian~Q Weinberger.
\newblock Densely connected convolutional networks.
\newblock In \emph{Proceedings of the IEEE conference on computer vision and
  pattern recognition}, pages 4700--4708, 2017{\natexlab{a}}.

\bibitem[Huang and Li(2021)]{huang2021mos}
Rui Huang and Yixuan Li.
\newblock Towards scaling out-of-distribution detection for large semantic
  space.
\newblock \emph{Proceedings of the IEEE/CVF Conference on Computer Vision and
  Pattern Recognition}, 2021.

\bibitem[Huang et~al.(2017{\natexlab{b}})Huang, Li, Poursaeed, Hopcroft, and
  Belongie]{huang2017stacked}
Xun Huang, Yixuan Li, Omid Poursaeed, John~E Hopcroft, and Serge~J Belongie.
\newblock Stacked generative adversarial networks.
\newblock In \emph{CVPR}, volume~2, page~3, 2017{\natexlab{b}}.

\bibitem[Kingma and Welling(2014)]{kingma2014autoencoding}
Diederik~P. Kingma and Max Welling.
\newblock Auto-encoding variational bayes.
\newblock In Yoshua Bengio and Yann LeCun, editors, \emph{2nd International
  Conference on Learning Representations, {ICLR} 2014}, 2014.

\bibitem[Kirichenko et~al.(2020)Kirichenko, Izmailov, and
  Wilson]{kirichenko2020normalizing}
Polina Kirichenko, Pavel Izmailov, and Andrew~G Wilson.
\newblock Why normalizing flows fail to detect out-of-distribution data.
\newblock \emph{Advances in Neural Information Processing Systems}, 33, 2020.

\bibitem[Kolesnikov et~al.(2020)Kolesnikov, Beyer, Zhai, Puigcerver, Yung,
  Gelly, and Houlsby]{kolesnikov2020big}
Alexander Kolesnikov, Lucas Beyer, Xiaohua Zhai, Joan Puigcerver, Jessica Yung,
  Sylvain Gelly, and Neil Houlsby.
\newblock Big transfer (bit): General visual representation learning.
\newblock In \emph{ECCV 2020}, 2020.

\bibitem[Krizhevsky et~al.(2009)Krizhevsky, Hinton,
  et~al.]{krizhevsky2009learning}
Alex Krizhevsky, Geoffrey Hinton, et~al.
\newblock Learning multiple layers of features from tiny images.
\newblock 2009.

\bibitem[Kullback and Leibler(1951)]{kullback1951information}
Solomon Kullback and Richard~A Leibler.
\newblock On information and sufficiency.
\newblock \emph{The annals of mathematical statistics}, 22\penalty0
  (1):\penalty0 79--86, 1951.

\bibitem[Lakshminarayanan et~al.(2017)Lakshminarayanan, Pritzel, and
  Blundell]{lakshminarayanan2017simple}
Balaji Lakshminarayanan, Alexander Pritzel, and Charles Blundell.
\newblock Simple and scalable predictive uncertainty estimation using deep
  ensembles.
\newblock In \emph{Advances in neural information processing systems}, pages
  6402--6413, 2017.

\bibitem[Lee and AlRegib(2020)]{lee2020gradients}
Jinsol Lee and Ghassan AlRegib.
\newblock Gradients as a measure of uncertainty in neural networks.
\newblock In \emph{2020 IEEE International Conference on Image Processing
  (ICIP)}, pages 2416--2420. IEEE, 2020.

\bibitem[Lee et~al.(2018)Lee, Lee, Lee, and Shin]{lee2018simple}
Kimin Lee, Kibok Lee, Honglak Lee, and Jinwoo Shin.
\newblock A simple unified framework for detecting out-of-distribution samples
  and adversarial attacks.
\newblock In \emph{Advances in Neural Information Processing Systems}, pages
  7167--7177, 2018.

\bibitem[Liang et~al.(2018)Liang, Li, and Srikant]{liang2018enhancing}
Shiyu Liang, Yixuan Li, and Rayadurgam Srikant.
\newblock Enhancing the reliability of out-of-distribution image detection in
  neural networks.
\newblock In \emph{6th International Conference on Learning Representations,
  ICLR 2018}, 2018.

\bibitem[Lin et~al.(2021)Lin, Roy, and Li]{lin2021mood}
Ziqian Lin, Sreya~Dutta Roy, and Yixuan Li.
\newblock Mood: Multi-level out-of-distribution detection.
\newblock In \emph{Proceedings of the IEEE/CVF Conference on Computer Vision
  and Pattern Recognition}, 2021.

\bibitem[Liu et~al.(2020)Liu, Wang, Owens, and Li]{liu2020energy}
Weitang Liu, Xiaoyun Wang, John Owens, and Yixuan Li.
\newblock Energy-based out-of-distribution detection.
\newblock \emph{Advances in Neural Information Processing Systems}, 2020.

\bibitem[Malinin et~al.(2021)Malinin, Band, Chesnokov, Gal, Gales, Noskov,
  Ploskonosov, Prokhorenkova, Provilkov, Raina, et~al.]{malinin2021shifts}
Andrey Malinin, Neil Band, German Chesnokov, Yarin Gal, Mark~JF Gales, Alexey
  Noskov, Andrey Ploskonosov, Liudmila Prokhorenkova, Ivan Provilkov, Vatsal
  Raina, et~al.
\newblock Shifts: A dataset of real distributional shift across multiple
  large-scale tasks.
\newblock \emph{arXiv preprint arXiv:2107.07455}, 2021.

\bibitem[Mohseni et~al.(2020)Mohseni, Pitale, Yadawa, and
  Wang]{mohseni2020self}
Sina Mohseni, Mandar Pitale, JBS Yadawa, and Zhangyang Wang.
\newblock Self-supervised learning for generalizable out-of-distribution
  detection.
\newblock In \emph{Proceedings of the AAAI Conference on Artificial
  Intelligence}, volume~34, pages 5216--5223, 2020.

\bibitem[Nalisnick et~al.(2018)Nalisnick, Matsukawa, Teh, Gorur, and
  Lakshminarayanan]{nalisnick2018deep}
Eric Nalisnick, Akihiro Matsukawa, Yee~Whye Teh, Dilan Gorur, and Balaji
  Lakshminarayanan.
\newblock Do deep generative models know what they don't know?
\newblock In \emph{International Conference on Learning Representations}, 2018.

\bibitem[Netzer et~al.(2011)Netzer, Wang, Coates, Bissacco, Wu, and
  Ng]{netzer2011reading}
Yuval Netzer, Tao Wang, Adam Coates, Alessandro Bissacco, Bo~Wu, and Andrew~Y
  Ng.
\newblock Reading digits in natural images with unsupervised feature learning.
\newblock 2011.

\bibitem[Nguyen et~al.(2015)Nguyen, Yosinski, and Clune]{nguyen2015deep}
Anh Nguyen, Jason Yosinski, and Jeff Clune.
\newblock Deep neural networks are easily fooled: High confidence predictions
  for unrecognizable images.
\newblock In \emph{Proceedings of the IEEE conference on computer vision and
  pattern recognition}, pages 427--436, 2015.

\bibitem[Oord et~al.(2016)Oord, Kalchbrenner, Vinyals, Espeholt, Graves, and
  Kavukcuoglu]{oord2016conditional}
A{\"a}ron van~den Oord, Nal Kalchbrenner, Oriol Vinyals, Lasse Espeholt, Alex
  Graves, and Koray Kavukcuoglu.
\newblock Conditional image generation with pixelcnn decoders.
\newblock In \emph{Proceedings of the 30th International Conference on Neural
  Information Processing Systems}, pages 4797--4805, 2016.

\bibitem[Ovadia et~al.(2019)Ovadia, Fertig, Ren, Nado, Sculley, Nowozin,
  Dillon, Lakshminarayanan, and Snoek]{ovadia2019can}
Yaniv Ovadia, Emily Fertig, Jie Ren, Zachary Nado, D~Sculley, Sebastian
  Nowozin, Joshua Dillon, Balaji Lakshminarayanan, and Jasper Snoek.
\newblock Can you trust your model's uncertainty? evaluating predictive
  uncertainty under dataset shift.
\newblock \emph{Advances in Neural Information Processing Systems},
  32:\penalty0 13991--14002, 2019.

\bibitem[Ren et~al.(2019)Ren, Liu, Fertig, Snoek, Poplin, Depristo, Dillon, and
  Lakshminarayanan]{ren2019likelihood}
Jie Ren, Peter~J Liu, Emily Fertig, Jasper Snoek, Ryan Poplin, Mark Depristo,
  Joshua Dillon, and Balaji Lakshminarayanan.
\newblock Likelihood ratios for out-of-distribution detection.
\newblock In \emph{Advances in Neural Information Processing Systems}, pages
  14680--14691, 2019.

\bibitem[Rezende et~al.(2014)Rezende, Mohamed, and
  Wierstra]{rezende2014stochastic}
Danilo~Jimenez Rezende, Shakir Mohamed, and Daan Wierstra.
\newblock Stochastic backpropagation and approximate inference in deep
  generative models.
\newblock In \emph{International conference on machine learning}, pages
  1278--1286. PMLR, 2014.

\bibitem[Schirrmeister et~al.(2020)Schirrmeister, Zhou, Ball, and
  Zhang]{schirrmeister2020understanding}
Robin~Tibor Schirrmeister, Yuxuan Zhou, Tonio Ball, and Dan Zhang.
\newblock Understanding anomaly detection with deep invertible networks through
  hierarchies of distributions and features.
\newblock \emph{arXiv preprint arXiv:2006.10848}, 2020.

\bibitem[Serrà et~al.(2020)Serrà, Álvarez, Gómez, Slizovskaia, Núñez, and
  Luque]{serra2019input}
Joan Serrà, David Álvarez, Vicenç Gómez, Olga Slizovskaia, José~F.
  Núñez, and Jordi Luque.
\newblock Input complexity and out-of-distribution detection with
  likelihood-based generative models.
\newblock In \emph{International Conference on Learning Representations}, 2020.

\bibitem[Tabak and Turner(2013)]{tabak2013family}
Esteban~G Tabak and Cristina~V Turner.
\newblock A family of nonparametric density estimation algorithms.
\newblock \emph{Communications on Pure and Applied Mathematics}, 66\penalty0
  (2):\penalty0 145--164, 2013.

\bibitem[Van~Horn et~al.(2018)Van~Horn, Mac~Aodha, Song, Cui, Sun, Shepard,
  Adam, Perona, and Belongie]{van2018inaturalist}
Grant Van~Horn, Oisin Mac~Aodha, Yang Song, Yin Cui, Chen Sun, Alex Shepard,
  Hartwig Adam, Pietro Perona, and Serge Belongie.
\newblock The inaturalist species classification and detection dataset.
\newblock In \emph{Proceedings of the IEEE conference on computer vision and
  pattern recognition}, pages 8769--8778, 2018.

\bibitem[Wang et~al.(2021)Wang, Liu, Bocchieri, and Li]{wang2021canmulti}
Haoran Wang, Weitang Liu, Alex Bocchieri, and Yixuan Li.
\newblock Can multi-label classification networks know what they don't know?
\newblock \emph{Advances in Neural Information Processing Systems}, 2021.

\bibitem[Wang et~al.(2020)Wang, Dai, Wipf, and Zhu]{wang2020further}
Ziyu Wang, Bin Dai, David Wipf, and Jun Zhu.
\newblock Further analysis of outlier detection with deep generative models.
\newblock \emph{Advances in Neural Information Processing Systems}, 33, 2020.

\bibitem[Winkens et~al.(2020)Winkens, Bunel, Roy, Stanforth, Natarajan, Ledsam,
  MacWilliams, Kohli, Karthikesalingam, Kohl, et~al.]{winkens2020contrastive}
Jim Winkens, Rudy Bunel, Abhijit~Guha Roy, Robert Stanforth, Vivek Natarajan,
  Joseph~R Ledsam, Patricia MacWilliams, Pushmeet Kohli, Alan Karthikesalingam,
  Simon Kohl, et~al.
\newblock Contrastive training for improved out-of-distribution detection.
\newblock \emph{arXiv preprint arXiv:2007.05566}, 2020.

\bibitem[Xiao et~al.(2010)Xiao, Hays, Ehinger, Oliva, and
  Torralba]{xiao2010sun}
Jianxiong Xiao, James Hays, Krista~A Ehinger, Aude Oliva, and Antonio Torralba.
\newblock Sun database: Large-scale scene recognition from abbey to zoo.
\newblock In \emph{2010 IEEE computer society conference on computer vision and
  pattern recognition}, pages 3485--3492. IEEE, 2010.

\bibitem[Xiao et~al.(2020)Xiao, Yan, and Amit]{xiao2020likelihood}
Zhisheng Xiao, Qing Yan, and Yali Amit.
\newblock Likelihood regret: An out-of-distribution detection score for
  variational auto-encoder.
\newblock \emph{Advances in Neural Information Processing Systems}, 33, 2020.

\bibitem[Yu et~al.(2015)Yu, Seff, Zhang, Song, Funkhouser, and
  Xiao]{yu2015lsun}
Fisher Yu, Ari Seff, Yinda Zhang, Shuran Song, Thomas Funkhouser, and Jianxiong
  Xiao.
\newblock Lsun: Construction of a large-scale image dataset using deep learning
  with humans in the loop.
\newblock \emph{arXiv preprint arXiv:1506.03365}, 2015.

\bibitem[Zhou et~al.(2017)Zhou, Lapedriza, Khosla, Oliva, and
  Torralba]{zhou2017places}
Bolei Zhou, Agata Lapedriza, Aditya Khosla, Aude Oliva, and Antonio Torralba.
\newblock Places: A 10 million image database for scene recognition.
\newblock \emph{IEEE transactions on pattern analysis and machine
  intelligence}, 40\penalty0 (6):\penalty0 1452--1464, 2017.

\end{thebibliography}
\bibliographystyle{plainnat}

\newpage
\appendix
\onecolumn
  
\begin{center}
    \Large{\textbf{Supplementary Material}}
\end{center}

\section{Evaluation on CIFAR Benchmarks}
\label{app:cifar}
\textbf{Setup}
We additionally evaluate \texttt{GradNorm} on a common benchmark with CIFAR-10 and CIFAR-100~\cite{krizhevsky2009learning} as ID datasets, which is routinely used in literature~\cite{hendrycks2016baseline,liang2018enhancing,hsu2020generalized,liu2020energy,lee2018simple}. We use the standard split with 50,000 training images and 10,000 test images. We pre-train a ResNet-20~\cite{he2016deep} network for 100 epochs. The learning rate is initially 0.1, and decays by a factor of 10 at epochs 50, 75 and 90 respectively. We evaluate on four common OOD benchmark datasets: \texttt{SVHN}~\cite{netzer2011reading}, \texttt{LSUN} (crop)~\cite{yu2015lsun}, \texttt{Places365}~\cite{zhou2017places}, and \texttt{Textures}~\cite{cimpoi14describing}. 

\textbf{Results} 
We summarize the results in Table~\ref{tab:cifar_results}, where \texttt{GradNorm} remains competitive. In particular, \texttt{GradNorm} reduces the average FPR95 by \textbf{8.77\%} on CIFAR-10 compared to the best baseline. On CIFAR-100, \texttt{GradNorm} outperforms the best baseline energy score~\cite{liu2020energy} by \textbf{14.47}\% in FPR95. Together with our large-scale evaluation in Section~\ref{sec:ablations}, \texttt{GradNorm} overall demonstrates superior performance compared to competitive methods in literature. While some baselines may require validation datasets, \texttt{GradNorm} is hyperparameter-free and can be used in OOD-agnostic setting. {Experimental details and hyperparameters of baseline methods can be found in Appendix~\ref{app:baseline}}. 

\begin{table}[h]
    \centering
    \scriptsize{

\begin{tabular}{c|l|C{0.03\textwidth}C{0.045\textwidth}|C{0.03\textwidth}C{0.045\textwidth}|C{0.03\textwidth}C{0.045\textwidth}|C{0.03\textwidth}C{0.045\textwidth}|C{0.03\textwidth}C{0.045\textwidth}}
\toprule
\multirow{3}{*}{\textbf{ID Data}} & \multicolumn{1}{c|}{\multirow{3}{*}{\textbf{Method}}} & \multicolumn{2}{c|}{\textbf{SVHN}}           & \multicolumn{2}{c|}{\textbf{LSUN (crop)}}    & \multicolumn{2}{c|}{\textbf{Places365}}      & \multicolumn{2}{c|}{\textbf{Textures}}       & \multicolumn{2}{c}{\textbf{Average}}        \\ \cline{3-12} 
                                & \multicolumn{1}{c|}{}                                 & \tiny{FPR95}                & \tiny{AUROC}                 & \tiny{FPR95}                & \tiny{AUROC}              & \tiny{FPR95}                & \tiny{AUROC}                 & \tiny{FPR95}                & \tiny{AUROC}                 & \tiny{FPR95}                & \tiny{AUROC}               \\
                                                                                      & \multicolumn{1}{c|}{}                                 & \multicolumn{1}{c}{$\downarrow$} & \multicolumn{1}{c|}{$\uparrow$} & \multicolumn{1}{c}{$\downarrow$} & \multicolumn{1}{c|}{$\uparrow$} & \multicolumn{1}{c}{$\downarrow$} & \multicolumn{1}{c|}{$\uparrow$} & \multicolumn{1}{c}{$\downarrow$} & \multicolumn{1}{c|}{$\uparrow$} & \multicolumn{1}{c}{$\downarrow$} & \multicolumn{1}{c}{$\uparrow$}  \\ \midrule
\multirow{5}{*}{CIFAR-10}         & MSP\tiny{~\cite{hendrycks2016baseline}}                                                   & 66.09                & 89.86                 & 37.73                & 94.87                 & 70.05                & 85.99                 & 68.23                & 87.62                 & 60.53                & 89.59                \\
                                  & ODIN\tiny{~\cite{liang2018enhancing}}                                                  & 55.52                & 89.63                 & 2.32                 & 99.39                 & {45.86}       & {90.81}        & 52.78                & 89.99                 & 39.12                & 92.46                \\
                                  & Energy\tiny{~\cite{liu2020energy}}                                                & 49.80                & 91.97                 & 3.86                 & 99.03                 & 46.48                & 90.55                 & 58.67                & 88.79                 & 39.70                & 92.59                \\
                                   & Mahalanobis\tiny{~\cite{lee2018simple}}                                           & 20.91 &	95.99 &	9.66 &	97.90 &	89.24 &	61.15 &	28.83 &	92.31 &	37.16 &	86.84              \\
                                  & \textbf{GradNorm (ours)}                              & {17.76}       & {96.66}        & {0.23}        & {99.87}        & 57.85                & 85.20                 & {37.71}       & {90.76}        & \textbf{28.39}       & \textbf{93.12}       \\ 
                                  \midrule
\multirow{5}{*}{CIFAR-100}        & MSP\tiny{~\cite{hendrycks2016baseline}}                                                   & 86.33                & 72.56                 & 66.33                & 82.06                 & 87.57                & 69.05                 & 90.64                & 64.02                 & 82.72                & 71.92                \\
                                  & ODIN\tiny{~\cite{liang2018enhancing}}                                                  & 94.80                & 66.85                 & 26.14                & 95.09                 & 82.57                & 72.90                 & 89.91                & 66.35                 & 73.36                & 75.30                \\
                                  & Energy\tiny{~\cite{liu2020energy}}                                                & 89.03                & 76.42                 & 21.90                & 95.90                 & {82.55}       & {72.98}        & 88.81                & 66.74                 & 70.57                & 78.01                \\
                                  & Mahalanobis\tiny{~\cite{lee2018simple}}                                           & 81.46 &	81.71 &	68.97 &	90.74 &	96.50 &	50.35 &	42.71 &	87.48 &	72.41 &	77.57                \\
                                  & \textbf{GradNorm (ours)}                              & 76.77                & 79.35                 & {1.12}        & {99.69}        & 88.74                & 65.99                 & {57.75}       & {81.83}        & \textbf{56.10}       & \textbf{81.72}       \\ 
\bottomrule
\end{tabular}
}
    \caption{\small{OOD detection performance on CIFAR-10 and CIFAR-100 benchmark. All methods utilize the standard ResNet-20~\cite{he2016deep} network.
}}
    \label{tab:cifar_results}
    \vspace{-0.5cm}
\end{table}

\section{Details of Experiments}
\label{app:exp_detail}

\subsection{Datasets}
\label{app:dataset}
\paragraph{Large-scale evaluation} We use ImageNet-1k~\cite{deng2009imagenet} as the in-distribution dataset, and evaluate on four OOD test datasets following the setup in \cite{huang2021mos}:
\begin{itemize}
    \item \textbf{iNaturalist}~\cite{van2018inaturalist} contains 859,000 plant and animal images across over 5,000 different species. Each image is resized to have a max dimension of 800 pixels. 
    We evaluate on 10,000 images randomly sampled from 110 classes that are disjoint from ImageNet-1k.
    \vspace{-0.1cm}
    \item \textbf{SUN}~\cite{xiao2010sun} contains over 130,000 images of scenes spanning 397 categories. SUN and ImageNet-1k have overlapping categories. 
    We evaluate on 10,000 images randomly sampled from 50 classes that are disjoint from ImageNet labels.
    \vspace{-0.1cm}
    \item \textbf{Places}~\cite{zhou2017places} is another scene dataset with similar concept coverage as SUN. A chosen subset of 10,000 images across 50 classes (not contained in ImageNet-1k) are used.
    \vspace{-0.1cm}
    \item \textbf{Textures}~\cite{cimpoi14describing} contains 5,640 real-world texture images under 47 categories. We use the entire dataset for evaluation.
\end{itemize}

\paragraph{CIFAR benchmark} CIFAR-10 and CIFAR-100~\cite{krizhevsky2009learning} are widely used as ID datasets in the literature, which contain 10 and 100 classes, respectively. We use the standard split with 50,000 training images and 10,000 test images. We evaluate our approach on four common OOD datasets, which are listed below:
\begin{itemize}
    \item \textbf{SVHN}~\cite{netzer2011reading} contains color images of house numbers. There are ten classes of digits 0-9. We use the entire test set containing 26,032 images.
    \vspace{-0.1cm}
    \item \textbf{LSUN}~\cite{yu2015lsun} contains 10,000 testing images across 10 different scenes. Image patches of size 32$\times$32 are randomly cropped from this dataset.
    \vspace{-0.1cm}
    \item \textbf{Places365}~\cite{zhou2017places} contains large-scale photographs of scenes with 365 scene categories. There are 900 images per category in the test set. We randomly sample 10,000 images from the test set for evaluation.
    \vspace{-0.1cm}
    \item \textbf{Textures}~\cite{cimpoi14describing} contains 5,640 real-world texture images under 47 categories. We use the entire dataset for evaluation.
\end{itemize}

\subsection{Details of Baselines}
\label{app:baseline}

For the reader's convenience, we summarize in detail a few common techniques for defining OOD scores that measure the degree of ID-ness on the given sample.
By convention, a higher (lower) score is indicative of being in-distribution (out-of-distribution).

\textbf{MSP~\cite{hendrycks2016baseline}} propose to use the maximum softmax score to detect OOD samples.

\paragraph{ODIN~\cite{liang2018enhancing}} \citeauthor{liang2018enhancing} improved OOD detection with temperature scaling and input perturbation. In all experiments, we set the temperature scaling parameter $T = 1000$. Note that this is different from calibration, where a much milder $T$ will be employed. While calibration focuses on representing the
true correctness likelihood of in-distribution data, the OOD scores proposed by ODIN are designed to maximize the gap between ID and OOD data and may no longer be meaningful from a predictive confidence standpoint.
For ImageNet, we found the input perturbation does not further improve the OOD detection performance and hence we set $\epsilon=0$. 
Following the setting in~\cite{liang2018enhancing}, we set $\epsilon$ to be 0.004 for CIFAR-10 and CIFAR-100.


\textbf{Energy~\cite{liu2020energy}} \citeauthor{liu2020energy} first proposed using energy score for OOD uncertainty estimation. The energy function  maps the logit outputs to a scalar $S_\mathrm{Energy}(\*x; f) \in \mathbb{R}$, which is relatively lower for ID data:
\begin{align}
\label{eq:energy}
S_\mathrm{Energy}(\*x; f)& =-\log \sum_{i=1}^{C}\exp(f_i(\*x)).
\end{align}
Note that \citeauthor{liu2020energy}~\cite{liu2020energy} used the \emph{negative energy score} for OOD detection, in order to align with the convention that $S(\*x;f)$ is higher (lower) for ID (OOD) data. Energy score is a hyperparameter-free score. 

\paragraph{Mahalanobis~\cite{lee2018simple}} \citeauthor{lee2018simple} use multivariate Gaussian distributions to model class-conditional distributions of softmax neural classifiers and use Mahalanobis distance-based scores for OOD detection. We use 500 examples randomly selected from ID datasets and an auxiliary tuning dataset to train the logistic regression model and tune the perturbation magnitude $\epsilon$. The tuning dataset consists of adversarial examples generated by FGSM~\cite{goodfellow2014explaining} with a perturbation size of 0.05.\@
The selected $\epsilon$'s are 0.001, 0.01, and 0.005 for ImageNet-1k, CIFAR-10, and CIFAR-100, respectively.


\subsection{Software and Hardware}
\paragraph{Software} We run all experiments with Python 3.8.0 and PyTorch 1.6.0.
\paragraph{Hardware} All experiments are run on NVIDIA GeForce RTX 2080Ti.

\section{Complete Results Under Different $L_p$-norms}
\label{app:more_norm}
In Table~\ref{tab:more_norm}, we report the OOD detection performance of using more $L_p$-norms as OOD scores, in addition to the 6 norms shown in Figure~\ref{fig:norm_ablation}. $L_1$-norm achieves the best overall performance.

\begin{table}[t]
    \centering
\small 
\begin{tabular}{c|c|cc}
\toprule
\multirow{2}{*}{\textbf{OOD Data}}     & \multirow{2}{*}{\textbf{Norm}} & \textbf{FPR95}       & \textbf{AUROC}       \\
                                       &                                & \multicolumn{1}{c}{$\downarrow$} & \multicolumn{1}{c}{$\uparrow$} \\ \midrule
\multirow{10}{*}{\textbf{iNaturalist}} & $L_{0.3}$                           & 60.30                & 83.45                \\
                                       & $L_{0.5}$                           & 65.28                & 81.44                \\
                                       & $L_{0.8}$                           & 66.25                & 82.69                \\
                                       & $L_{1}$                             & \textbf{50.03}       & \textbf{90.33}       \\
                                       & $L_{2}$                             & 65.88                & 87.68                \\
                                       & $L_{3}$                             & 71.15                & 85.71                \\
                                       & $L_{4}$                             & 73.53                & 84.67                \\
                                       & $L_{5}$                             & 74.66                & 84.13                \\
                                       & $L_{6}$                             & 75.19                & 83.83                \\
                                       & $L_{\infty}$                           & 76.42                & 83.18                \\ \midrule
\multirow{10}{*}{\textbf{SUN}}         & $L_{0.3}$                           & 48.53                & 86.73                \\
                                       & $L_{0.5}$                           & 51.98                & 85.13                \\
                                       & $L_{0.8}$                           & 54.52                & 84.75                \\
                                       & $L_{1}$                             & \textbf{46.48}       & \textbf{89.03}       \\
                                       & $L_{2}$                             & 76.09                & 80.30                \\
                                       & $L_{3}$                             & 80.87                & 77.85                \\
                                       & $L_{4}$                             & 82.59                & 76.65                \\
                                       & $L_{5}$                             & 83.37                & 76.01                \\
                                       & $L_{6}$                             & 83.73                & 75.64                \\
                                       & $L_{\infty}$                           & 84.41                & 74.83                \\ \midrule
\multirow{10}{*}{\textbf{Places}}      & $L_{0.3}$                          & 64.10                & 81.46                \\
                                       & $L_{0.5}$                           & 67.39                & 79.79                \\
                                       & $L_{0.8}$                          & 69.01                & 79.83                \\
                                       & $L_{1}$                             & \textbf{60.86}       & \textbf{84.82}       \\
                                       & $L_{2}$                             & 80.14                & 77.80                \\
                                       & $L_{3}$                             & 82.68                & 75.69                \\
                                       & $L_{4}$                             & 83.70                & 74.61                \\
                                       & $L_{5}$                             & 84.24                & 74.03                \\
                                       & $L_{6}$                             & 84.34                & 73.70                \\
                                       & $L_{\infty}$                           & 85.01                & 72.95                \\ \midrule
\multirow{10}{*}{\textbf{Textures}}    & $L_{0.3}$                           & 53.21                & 82.82                \\
                                       & $L_{0.5}$                           & 41.54                & 88.46                \\
                                       & $L_{0.8}$                          & \textbf{33.40}       & \textbf{92.07}       \\
                                       & $L_{1}$                             & 61.42                & 81.07                \\
                                       & $L_{2}$                             & 85.20                & 70.21                \\
                                       & $L_{3}$                             & 87.50                & 67.54                \\
                                       & $L_{4}$                             & 88.30                & 66.68                \\
                                       & $L_{5}$                             & 88.49                & 66.40                \\
                                       & $L_{6}$                             & 88.37                & 66.29                \\
                                       & $L_{\infty}$                           & 88.40                & 66.13                \\ \midrule
\multirow{10}{*}{\textbf{Average}}     & $L_{0.3}$                           & 56.54                & 83.62                \\
                                       & $L_{0.5}$                           & 56.55                & 83.71                \\
                                       & $L_{0.8}$                          & 55.80                & 84.84                \\
                                       & $L_{1}$                             & \textbf{54.70}       & \textbf{86.31}       \\
                                       & $L_{2}$                             & 76.83                & 79.00                \\
                                       & $L_{3}$                             & 80.55                & 76.70                \\
                                       & $L_{4}$                             & 82.03                & 75.65                \\
                                       & $L_{5}$                             & 82.69                & 75.14                \\
                                       & $L_{6}$                             & 82.91                & 74.87                \\
                                       & $L_{\infty}$                           & 83.56                & 74.27                \\ \bottomrule
\end{tabular}
    \caption{OOD detection performance comparison under different $L_p$-norms. We use a ResNetv2-101 architecture pre-trained on ImageNet-1k.}
    \label{tab:more_norm}
\end{table}

\clearpage
\section{Complete Results Under Different Scaling Temperatures}
\label{app:more_temper}

In addition to Figure~\ref{fig:temper_ablation}, we report the OOD detection performance under more scaling temperatures in Table~\ref{tab:more_temper}. $T=1$ achieves the best average performance.

\begin{table}[h]
    \centering
    \scriptsize{
\begin{tabular}{c|C{0.04\textwidth}C{0.05\textwidth}|C{0.04\textwidth}C{0.05\textwidth}|C{0.04\textwidth}C{0.05\textwidth}|C{0.04\textwidth}C{0.05\textwidth}|C{0.04\textwidth}C{0.05\textwidth}}
\toprule
\multirow{3}{*}{\textbf{Temperature}} & \multicolumn{2}{c|}{\textbf{iNaturalist}}    & \multicolumn{2}{c|}{\textbf{SUN}}            & \multicolumn{2}{c|}{\textbf{Places}}         & \multicolumn{2}{c|}{\textbf{Textures}}       & \multicolumn{2}{c}{\textbf{Average}}        \\ \cline{2-11} 
                                      & \textbf{FPR95}       & \textbf{AUROC}        & \textbf{FPR95}       & \textbf{AUROC}        & \textbf{FPR95}       & \textbf{AUROC}        & \textbf{FPR95}       & \textbf{AUROC}        & \textbf{FPR95}       & \textbf{AUROC}       \\
                                      & \multicolumn{1}{c}{$\downarrow$} & \multicolumn{1}{c|}{$\uparrow$} & \multicolumn{1}{c}{$\downarrow$} & \multicolumn{1}{c|}{$\uparrow$} & \multicolumn{1}{c}{$\downarrow$} & \multicolumn{1}{c|}{$\uparrow$} & \multicolumn{1}{c}{$\downarrow$} & \multicolumn{1}{c|}{$\uparrow$} & \multicolumn{1}{c}{$\downarrow$} & \multicolumn{1}{c}{$\uparrow$} \\ \midrule
0.0625                                & 77.70                & 74.46                 & 61.87                & 78.80                 & 76.42                & 72.84                 & 67.80                & 72.84                 & 70.95                & 74.74                \\
0.125                                 & 77.47                & 74.68                 & 61.84                & 78.90                 & 76.35                & 72.97                 & 67.73                & 72.93                 & 70.85                & 74.87                \\
0.25                                  & 76.64                & 75.55                 & 61.52                & 79.25                 & 75.95                & 73.48                 & 67.41                & 73.30                 & 70.38                & 75.40                \\
0.5                                   & 70.80                & 80.01                 & 58.83                & 81.47                 & 72.95                & 76.38                 & 65.30                & 75.29                 & 66.97                & 78.29                \\
1                                     & \textbf{50.03}       & \textbf{90.33}        & 46.48                & 89.03                 & 60.86                & 84.82                 & \textbf{61.42}       & \textbf{81.07}        & \textbf{54.70}       & \textbf{86.31}       \\
2                                     & 58.06                & 87.33                 & \textbf{41.03}       & \textbf{91.67}        & \textbf{55.71}       & \textbf{87.54}        & 65.94                & 77.98                 & 55.19                & 86.13                \\
4                                     & 71.69                & 79.68                 & 47.37                & 89.87                 & 61.96                & 85.97                 & 71.21                & 70.96                 & 63.06                & 81.62                \\
8                                     & 77.47                & 76.01                 & 52.85                & 87.87                 & 66.76                & 83.82                 & 73.62                & 67.50                 & 67.68                & 78.80                \\
16                                    & 79.54                & 74.71                 & 55.08                & 86.93                 & 68.69                & 82.80                 & 74.73                & 66.24                 & 69.51                & 77.67                \\
32                                    & 80.25                & 74.18                 & 56.22                & 86.49                 & 69.81                & 82.33                 & 75.02                & 65.71                 & 70.33                & 77.18                \\
64                                    & 80.53                & 73.94                 & 56.53                & 86.28                 & 70.08                & 82.11                 & 75.09                & 65.48                 & 70.56                & 76.95                \\
128                                   & 80.70                & 73.83                 & 56.78                & 86.18                 & 70.24                & 82.00                 & 75.11                & 65.36                 & 70.71                & 76.84                \\
256                                   & 80.81                & 73.77                 & 56.94                & 86.13                 & 70.32                & 81.94                 & 75.12                & 65.31                 & 70.80                & 76.79                \\
512                                   & 80.84                & 73.74                 & 57.01                & 86.10                 & 70.39                & 81.91                 & 75.16                & 65.28                 & 70.85                & 76.76                \\
1024                                  & 80.85                & 73.73                 & 57.05                & 86.09                 & 70.41                & 81.90                 & 75.16                & 65.27                 & 70.87                & 76.75                \\ \bottomrule
\end{tabular}
}
    \caption{OOD detection performance comparison under different temperatures. We use a ResNetv2-101 architecture pre-trained on ImageNet-1k.}
    \label{tab:more_temper}
\end{table}

\section{Comparison with \citeauthor{lee2020gradients}}
\label{app:comparison_with_la}

\citeauthor{lee2020gradients}~\cite{lee2020gradients} proposed an OOD detection framework using the $L_2$-norm of gradients. Importantly, they do not directly use gradient norms for OOD detection, but instead, use them as the input for training a separate binary classifier. The binary classifier is trained on layer-wise gradients from both ID and OOD data. Since \texttt{GradNorm} does not require any OOD data, these two methods are not directly comparable. For a fair comparison, we use the gradients of uniform noise as a surrogate of OOD data to train the binary classifier. Following the original work, we use a 40\%-40\%-20\% train-validation-test split. 
Results are shown in Table~\ref{table:compare_with_lee_alregib}. \texttt{GradNorm} outperforms~\cite{lee2020gradients} by 15.45\% in FPR95 on average.

\begin{table*}[h]
    \centering
    \scriptsize{
\begin{tabular}{c|C{0.04\textwidth}C{0.055\textwidth}|C{0.04\textwidth}C{0.055\textwidth}|C{0.04\textwidth}C{0.055\textwidth}|C{0.04\textwidth}C{0.055\textwidth}|C{0.04\textwidth}C{0.055\textwidth}}
\toprule
\multirow{3}{*}{\textbf{Method}}              & \multicolumn{2}{c|}{\textbf{iNaturalist}}    & \multicolumn{2}{c|}{\textbf{SUN}}            & \multicolumn{2}{c|}{\textbf{Places}}         & \multicolumn{2}{c|}{\textbf{Textures}}       & \multicolumn{2}{c}{\textbf{Average}}        \\ \cline{2-11} 
                                              & FPR95                & AUROC                 & FPR95                & AUROC                 & FPR95                & AUROC                 & FPR95                & AUROC                 & FPR95                & AUROC                \\
                                              & \multicolumn{1}{c}{$\downarrow$} & \multicolumn{1}{c|}{$\uparrow$} & \multicolumn{1}{c}{$\downarrow$} & \multicolumn{1}{c|}{$\uparrow$} & \multicolumn{1}{c}{$\downarrow$} & \multicolumn{1}{c|}{$\uparrow$} & \multicolumn{1}{c}{$\downarrow$} & \multicolumn{1}{c|}{$\uparrow$} & \multicolumn{1}{c}{$\downarrow$} & \multicolumn{1}{c}{$\uparrow$} \\ \midrule
\multicolumn{1}{l|}{\citeauthor{lee2020gradients}\tiny{~\cite{lee2020gradients}}}           & 75.49                & 72.30                 & 73.82                & 82.61                 & 87.90                & 74.00                 & \textbf{43.39}       & \textbf{84.16}        & 70.15                & 78.27                \\
\multicolumn{1}{l|}{\textbf{GradNorm (ours)}} & \textbf{50.03}       & \textbf{90.33}        & \textbf{46.48}       & \textbf{89.03}        & \textbf{60.86}       & \textbf{84.82}        & 61.42                & 81.07                 & \textbf{54.70}       & \textbf{86.31}       \\ \bottomrule
\end{tabular}
}
\caption{\small{Comparison of \texttt{GradNorm} \citeauthor{lee2020gradients}'s approach on ImageNet benchmark. The classification model is the same as in Table~\ref{table:main} (standard ResNetv2-101 model pre-trained on ImageNet). For \citeauthor{lee2020gradients}'s method, we use the gradients of uniform noise as a surrogate of OOD data to train the binary classifier.}}
    \label{table:compare_with_lee_alregib}
\end{table*}

\end{document}